\documentclass[preprint,12pt]{elsarticle}

\usepackage{amssymb}
\usepackage{amsmath}
\usepackage{amsfonts}
\usepackage{enumitem}
\usepackage{algorithmic}
\usepackage{graphicx}
\usepackage{textcomp}
\usepackage{dsfont}
\usepackage{array}
\usepackage{multirow}
\usepackage{bm}
\usepackage{url}
\usepackage{pifont}
\usepackage{amssymb}
\usepackage{pifont}
\usepackage{tabularx}
\usepackage{makecell}
\usepackage{colortbl} 
\newcommand{\figref}[1]{Figure \ref{#1}}
\newcommand{\tabref}[1]{Table \ref{#1}}

\usepackage{epstopdf}
\usepackage[normalem]{ulem}
\useunder{\uline}{\ul}{}
\usepackage{booktabs} 
\usepackage[dvipsnames, table]{xcolor} 
\newcolumntype{C}{>{\centering\arraybackslash}m{1.6cm}}
\newcolumntype{M}[1]{>{\centering\arraybackslash}m{#1}}
\definecolor{mygreen}{RGB}{112,173,71}
\definecolor{myorange}{RGB}{244,177,131}

\usepackage{subcaption}

\journal{Information Fusion}

\begin{document}

\begin{frontmatter}

\title{\textbf{Text-Aided Multi-Modal Panoptic Symbol Spotting for CAD Floor Plan Drawings}}

\author[inst1]{Yan Gong\fnref{fn1}}
\author[inst1]{Bohao Li\fnref{fn1}}
\author[inst1,inst2]{Bowen Du}
\author[inst3]{Junchen Ye}

\fntext[fn1]{Equal contribution.}

\affiliation[inst1]{
    organization={CCSE Lab, Beihang University},
    city={Beijing},
    postcode={100191},
    country={China}
}

\affiliation[inst2]{
    organization={School of Transportation Science and Engineering, Beihang University},
    city={Beijing},
    postcode={100191},
    country={China}
}

\affiliation[inst3]{
    organization={The Hong Kong Polytechnic University},
    city={Hong Kong},
    country={China}
}

\begin{abstract}
Computer-Aided Design (CAD) floor plan drawings contain both graphical primitives and textual annotations, which provide complementary geometric and semantic cues for intelligent design understanding.
Among CAD analysis tasks, panoptic symbol spotting has become increasingly important with the growing demand for industrial digitalization and deep learning-based automation.
However, most existing methods remain primarily primitive-centric and underexploit textual annotations, despite their critical semantic value. 
Even the few text-aware approaches often treat annotations only superficially, without properly modeling complex syntax and hierarchical semantics of CAD annotations, which leads to semantic loss and suboptimal spotting performance.
To address these limitations, we propose TextCAD, a multimodal framework that jointly models graphical primitives and textual annotations for panoptic symbol spotting. 
Specifically, we design a Type--Attribute Correlation Encoder (TACE) to explicitly encode the compositional semantics within annotations by jointly modeling their types and attributes. 
We further introduce a Semantic Hierarchy Alignment framework with Multi-level Semantic Filtering (MSF) and primitive downsampling, which adaptively aligns annotation semantics with graphical primitives at different semantic levels and enables accurate cross-modal semantic injection and fusion.
Experiments on real-world building-design datasets show that TextCAD effectively improves symbol spotting performance and achieves state-of-the-art results.
\end{abstract}




\end{frontmatter}

\section{Introduction}
\label{sec:intro}

Computer-Aided Design (CAD) floor plans are vector-based design documents composed of fine-grained graphical primitives such as lines and arcs, which explicitly encode the structural details of buildings~\cite{hunde2022future,rezvanifar2019symbol}. Owing to their precise geometric expressiveness, CAD drawings have been widely used throughout Architecture, Engineering, and Construction (AEC) workflows~\cite{shivegowda2022review,zhao2017scientometric}. 

As intelligent Building Information Modeling (BIM) applications continue to advance, these drawings increasingly need to be parsed into structured semantic elements that support downstream understanding, reasoning, and digital management~\cite{gao2019bim,heidari2024systematic,yang2020semiautomatic}. 
In this context, \textit{panoptic symbol spotting} has emerged as a fundamental task for CAD analysis, aiming to jointly detect and classify all symbols in a floor plan, including both countable objects (e.g., doors and windows) and stuff-like structures (e.g., walls), at the primitive level~\cite{fan2021floorplancad}. It therefore serves as a key bridge between low-level CAD primitives and high-level BIM semantics~\cite{liu2024comparison}.

Existing panoptic symbol spotting methods have mainly focused on improving the representation of graphical primitives. Early studies rasterize CAD vectors into images and apply image-based recognition pipelines~\cite{fan2021floorplancad,pang2024pixel,rezvanifar2020symbol}, but rasterization often destroys the geometric and topological details that are critical in CAD floor plan drawings~\cite{crommelinck2016review}. Later methods instead operate directly on vector primitives and organize them as basic units in different network architectures, such as graph-based, transformer-based, point-based, or line-based models~\cite{jiang2021recognizing,zheng2022gat,yang2023vectorfloorseg,fan2022cadtransformer,liu2024symbol,liu2024sympoint,yang2024cadspotting,luo2026archcad,wei2026point}. Although these paradigms differ in representation and architecture, their common goal is to learn more discriminative primitive-level geometric features, and most of them remain predominantly primitive-centric.

\begin{figure}[tbp]
    \centering

    \begin{subfigure}[c]{0.4\linewidth}
        \centering
        \includegraphics[width=\linewidth]{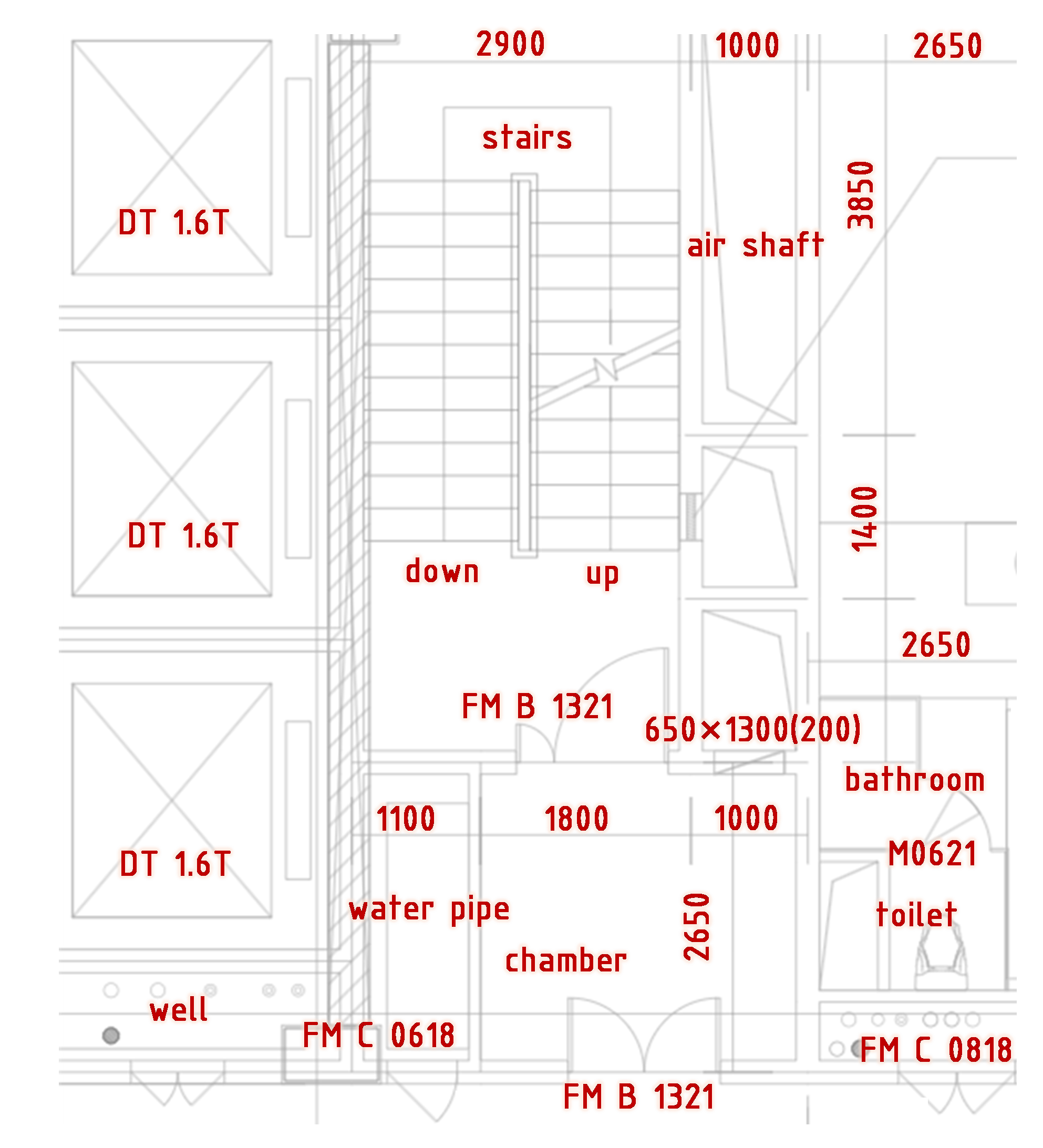}
        \caption{CAD with dense text notations.}
        \label{fig1:a}
    \end{subfigure}
    \begin{subfigure}[c]{0.4\linewidth}
        \centering

        \begin{subfigure}[b]{\linewidth}
            \centering
            \includegraphics[width=0.8\linewidth]{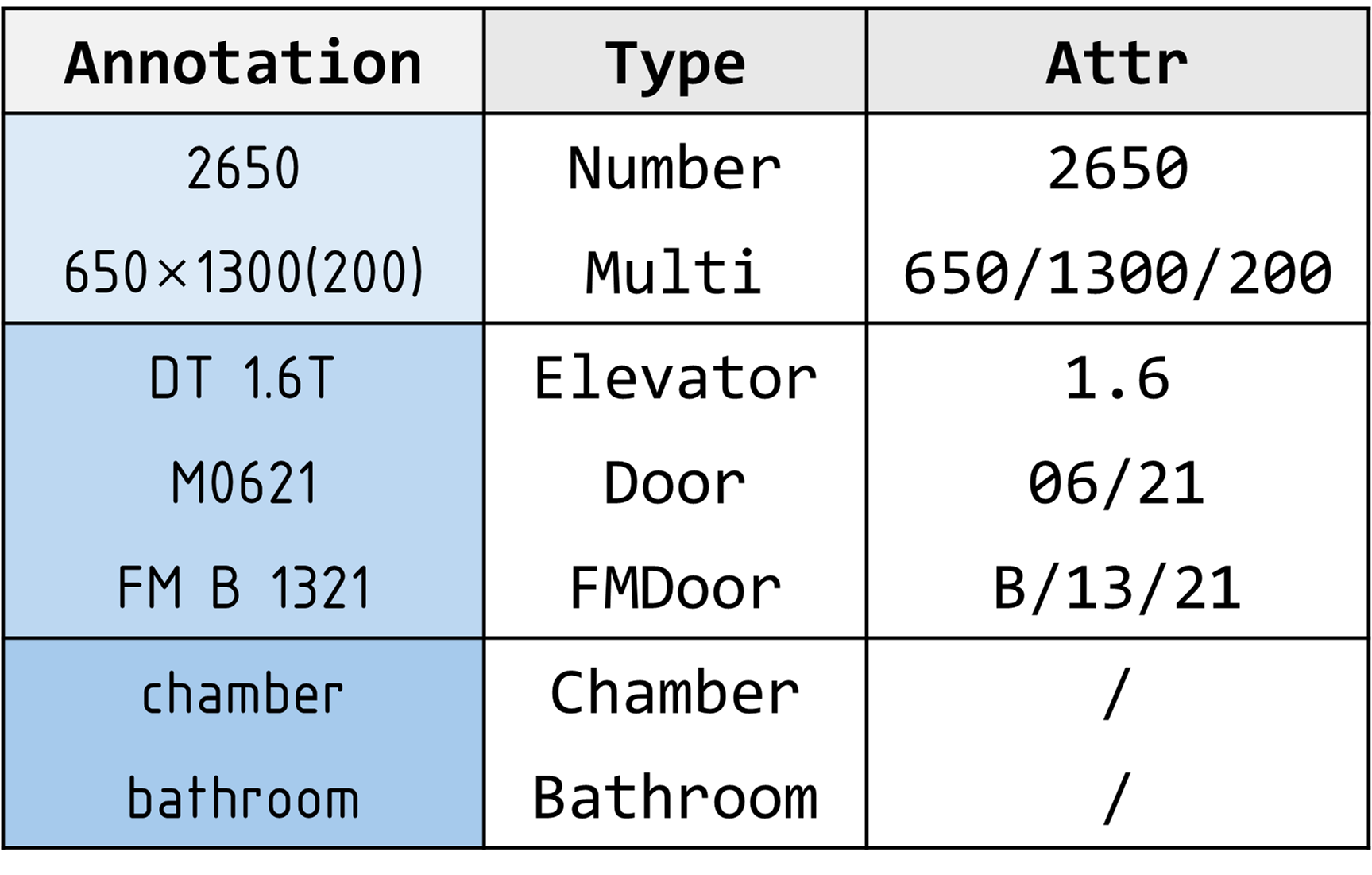}
            \caption{Complex semantic structure.}
            \label{fig1:b}
        \end{subfigure}

        \vspace{1mm}

        \begin{subfigure}[b]{\linewidth}
            \centering
            \includegraphics[width=0.8\linewidth]{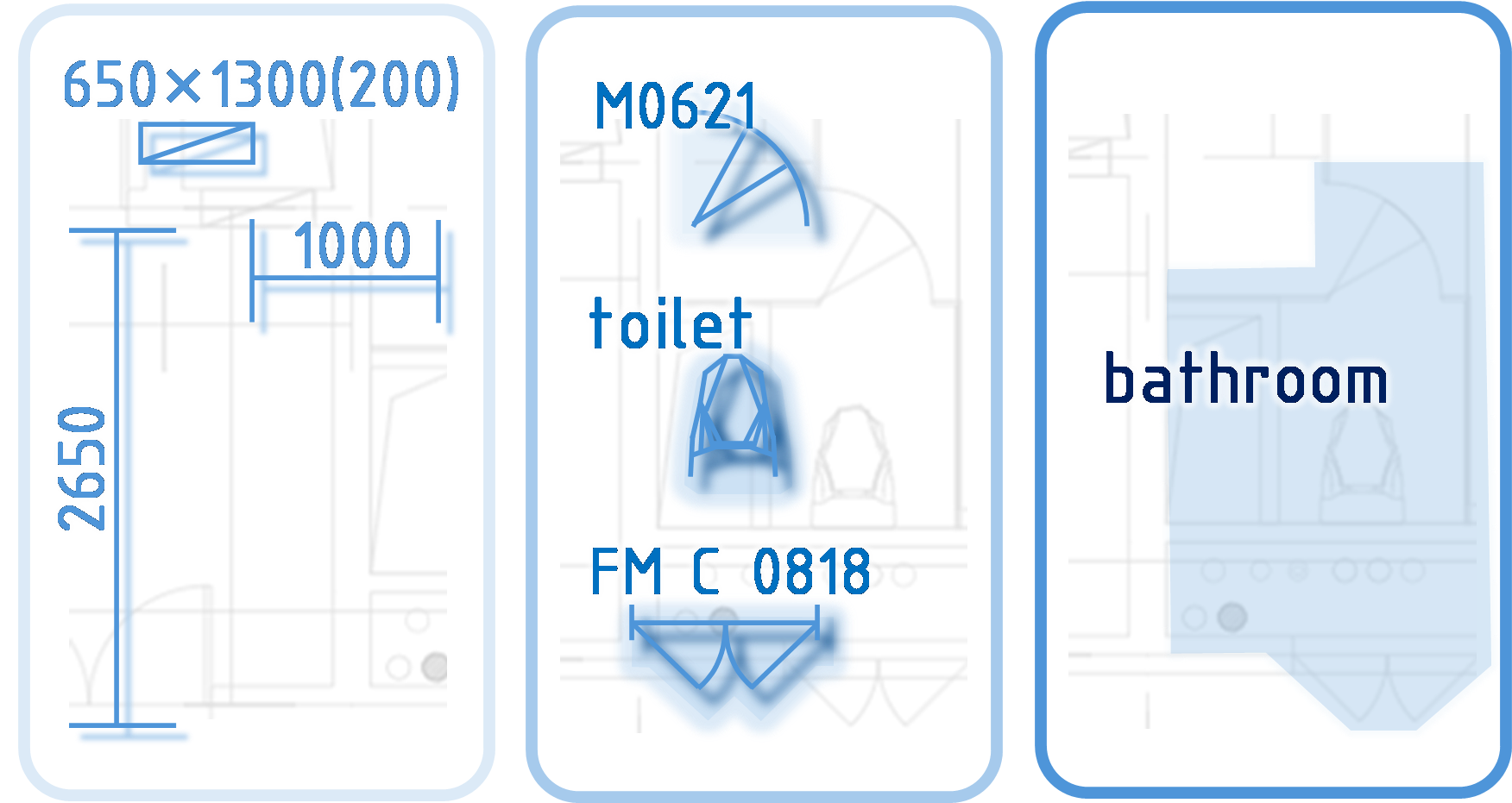}
            \caption{Multi-level semantic.}
            \label{fig1:c}
        \end{subfigure}
    \end{subfigure}

    \caption{Characteristics of CAD text annotations: (a) CAD drawings contain dense text annotations which convey rich semantic cues. In (b) and (c), we illustrate the complex syntactic structures and multi-level semantics of CAD textual annotations.}
    \label{fig1}
\end{figure}

However, CAD floor plan drawings are not purely geometric. 
In practice, they also contain \textit{dense textual annotations} that provide direct semantic cues about component identity, dimensions, functions, and material properties, as illustrated in \figref{fig1:a}. 
These annotations are often decisive for distinguishing structures that may be geometrically ambiguous, partially incomplete, or heavily overlapped. 
\textit{Therefore, the central problem is not merely how to improve geometric representation of primitives, but how to effectively incorporate CAD textual semantics into primitive-level spotting.}

Achieving this is nontrivial because CAD annotations are neither generic natural language nor flat auxiliary labels; rather, they exhibit \textit{complex syntactic structures} and \textit{multi-level semantics}. Consequently, existing attempts that rely on naive text encoding~\cite{wang2025panoramic} or direct text-to-primitive fusion~\cite{xing2026multimodal} may lose critical semantics. 
Accordingly, integrating textual annotations into panoptic symbol spotting still faces two key challenges:
1) \textbf{\textit{Concise yet dense CAD annotation syntax hinders unified type–attribute representation.}}
CAD text annotations use highly compressed symbolic codes to convey rich meaning.
For example, as shown in~\figref{fig1:b}, in the string `\texttt{FM B 1321}', `\texttt{FM}' denotes the type (fire door), while `\texttt{B 1321}' encodes two attributes—grade (Class B) and dimensions (13 dm × 21 dm).
These short strings carry high information density.
Off-the-shelf language models (LMs~\cite{devlin2019bert} / LLMs~\cite{brown2020language}) struggle here: their tokenizers often fail to segment such domain-specific patterns reliably~\cite{wang2025panoramic} and incur computational cost~\cite{zhu2024survey}. 
Simple alternatives (e.g. MLPs~\cite{rosenblatt1958perceptron} or lookup embeddings~\cite{mikolov2013efficient}) also fall short, as they cannot explicitly model and link type and attribute semantics, leading to representation bias.
Hence, a key challenge is to efficiently fuse type and attribute within annotations under lightweight computation to yield robust semantic representations.
2) \textbf{\textit{Multi-level semantics of CAD annotations make fusion with fine-grained primitives prone to semantic loss.}} 
As shown in ~\figref{fig1:c}, CAD annotations span multiple semantic scales: from the primitive-level (e.g. wall length `\texttt{2650}'), to the instance-level (e.g. `\texttt{M0621}'), and up to the region-level (e.g. `\texttt{bathroom}'). Different annotations have varying scopes of semantic coverage, corresponding to primitives at different geometric scales.
Naively injecting high-level annotations into low-level primitive representations can cause a mismatch between semantic scope and geometric receptive field; 
for example, attaching a region-level tag to a single local primitive can attenuate or mislocalize its information through message passing.
Therefore, a key challenge is to ensure that annotations operate at their corresponding semantic scales, avoiding information loss and cross-level interference.

To address the above challenges, we propose a novel panoptic symbol spotting model, \textbf{TextCAD}, which enables the effective embedding and fusion of textual annotations within CAD floor plan drawings.
Specifically, the model first embeds graphical primitives and textual annotations separately. For textual annotations, TextCAD introduces a Type-Attribute Correlation Encoder (TACE) to model the semantic correlations between annotation types and their attributes, thereby producing comprehensive textual representations. For multimodal fusion, TextCAD adopts a multi-level downsampling–upsampling architecture to capture design semantics and structural details at different levels. During downsampling, as primitive representations are progressively reduced and evolve to encode broader semantic context, the proposed Multi-level Semantic Filtering (MSF) module selects semantically corresponding textual features for the downsampled primitives and fuses them to achieve cross-modal semantic alignment across different levels. The upsampling stage then performs a symmetric restoration process. Finally, the refined primitive features are fed into a decoder to produce the final spotting results.
Extensive experiments on real-world building-design datasets show that TextCAD consistently outperforms existing baselines, especially in the absence of priors (such as layers).
Our main contributions are summarized as follows:
\begin{itemize}[leftmargin=*,itemsep=2pt, parsep=0pt]

\item \textit{Type-Attribute Correlation Encoder} is proposed to capture the compositional semantics of textual annotations. It formulates annotations under a unified \textit{type}--\textit{attribute} schema and models their semantic dependencies through attention-based interactions, thereby producing expressive semantic representations.
\item \textit{Multi-level Semantic Filtering (MSF)} is proposed to achieve level-consistent cross-modal fusion. Guided by downsampled primitive representations, MSF selectively injects hierarchy-consistent textual semantics into primitives at different stages, thereby reducing cross-level semantic mismatch and enabling semantically aligned fusion.
\item \textit{A multimodal panoptic symbol spotting framework} is proposed to integrate textual annotations with graphical primitives through a down--upsampling architecture. Experiments on real-world datasets show that it consistently improves the accuracy and robustness of symbol spotting over existing baselines, especially when prior information is unavailable.
\end{itemize}

\section{Preliminary}
\label{sec:preliminary}

This section formalizes the research task and notation for CAD floor plan drawings, covering primitives and text annotations.

\subsection{Problem formalization}
Given a CAD floor plan drawing, we consider two fundamental modalities: 
the graphical primitives $\mathbf{E}_g$ (including basic geometric elements such as lines, arcs, circles, and ellipses) and the text annotations $\mathbf{E}_t$.
We formalize our target panoptic symbol spotting task as:
\begin{equation}
f_{\boldsymbol{\theta}}:\; (\mathbf{E}_g, \mathbf{E}_t)\;\longrightarrow\; (\hat{\mathbf{Y}}, \hat{\mathbf{Z}}),
\end{equation}
where $\hat{\mathbf{Y}}=\{\hat y_i\}_{i=1}^{N_g}$ are the predicted semantic labels for primitives $e_g^{\,i}\in\mathbf{E}_g$, and $\hat{\mathbf{Z}}=\{\hat z_i\}_{i=1}^{N_g}$ are the corresponding instance indices. 
Here, $N_g = |\mathbf{E}_g|$ is the number of graphical primitives and $\boldsymbol{\theta}$ denotes the learnable model parameters. 
Each $\hat z_i \in \mathbb{Z}_{\ge 0}\cup\{-1\}$, with $\hat z_i=-1$ reserved for primitives without a specific instance~\cite{fan2021floorplancad}.

\subsection{Text annotations decomposition}
For each text annotation $e_t^{i} \in \mathbf{E}_t$, we write $e_t^i = \{ \mathcal{T}_i, \mathcal{A}_i \}$, where $\mathcal{T}_i$ represents the annotation-type indicator and $\mathcal{A}_i \in \mathbb{R}^a$ is the attribute vector separated from the raw annotation.
Here, $a$ denotes the maximum number of properties (e.g. length or grade) captured in the syntactic structures.
We additionally attach a mask vector $\mathcal{M}_i \in \mathbb{R}^a$ to indicate the validity of each attribute. 
The detailed procedure of attribute separation in \ref{sec:tace-decompose}.

\section{Methodology: TextCAD}
\label{sec:method}

\begin{figure*}[htbp]
  \centering
   \includegraphics[width=\linewidth]{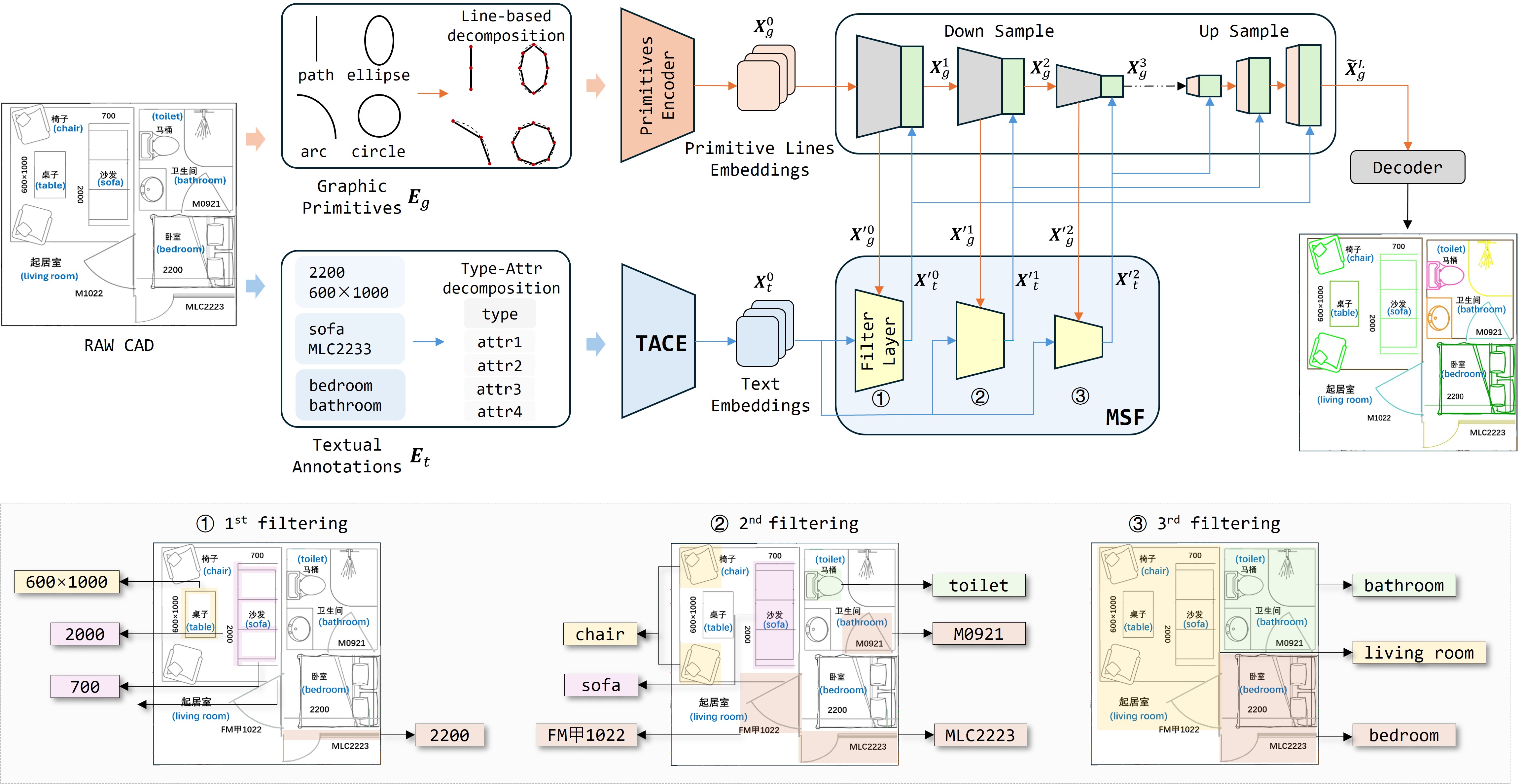}
   \caption{Framework of our method. TextCAD proposes Type-Attribute Correlation Encoder to embed textual annotations and employ down-upsampling architecture for modal fusion by designing multi-level semantic filtering for modal alignment.}
   \label{fig2}
\end{figure*}

In this section, we describe the proposed TextCAD.
As illustrated in ~\figref{fig2}, the workflow of our model is as follows: 

For a given CAD drawing, we first decompose it into a set of graphical primitives $\mathbf{E}_g$ and a set of text annotations $\mathbf{E}_t$, which are respectively embedded by unimodal encoders. Here, the \textit{Type-Attribute Correlation Encoder} (TACE) is employed to embed the text annotations information, capturing their semantics by jointly modeling type and attributes,
while graphical primitives are embedded based on the line-based method following ~\cite{wei2026point}.

Then we adopt down–upsampling architecture for multimodal fusion. As primitive features are aggregated through a downsampling process, textual features are progressively filtered through the \textit{Multi-Level Semantic Filtering} (MSF) mechanism to align the semantic hierarchy.
As shown at the bottom of ~\figref{fig2}, low-level, instance-level and high-level annotations are progressively aligned with the discrete, instance-level, and region-level representations of primitives respectively.
Note that this three-stage filtering is only a schematic illustration; 
in practice, MSF mechanism performs adaptive semantic alignment.
The filtered aligned text semantics are then fused with corresponding primitive features at multi-levels and go through the upsampling phase for feature restoration. 
Finally, the refined primitive features are passed to the decoder for final prediction.

\subsection{Unimodal encoder}

We first employ two modality-specific encoders to separately embed the graphical primitives and text annotations.

\subsubsection{Type-Attribute Correlation Encoder.}
In CAD floor plan drawings, text annotations are written in highly compressed symbolic forms that specify a \textit{type} and attach multiple \textit{attributes}. 
The \textit{type} conveys functional semantics, while the \textit{attributes} encode characteristics such as scope or capacity; both are critical cues. 
Moreover, different type–attribute patterns induce different attribute semantics. For example, the same numeric attribute string may denote load capacity when associated with \texttt{Elevator}, but dimensions when associated with \texttt{Door}.
Conventional NLP embeddings struggle in this high-semantic-density setting: language-model tokenizers are unstable, and simple embedding schemes cannot model the diverse type–attribute patterns. 

Therefore, we propose the Type–Attribute Correlation Encoder (TACE), which formulates CAD text annotations under a unified type–attribute representation schema and embeds them by leveraging cross type–attribute attention to associate each \textit{type} with its \textit{attribute} bundle, thereby capturing the essential semantics.
The robustness of TACE is discussed in \ref{sec:tace-extensibility}.

\begin{figure}[tbp]
  \centering
   \includegraphics[width=0.8\linewidth]{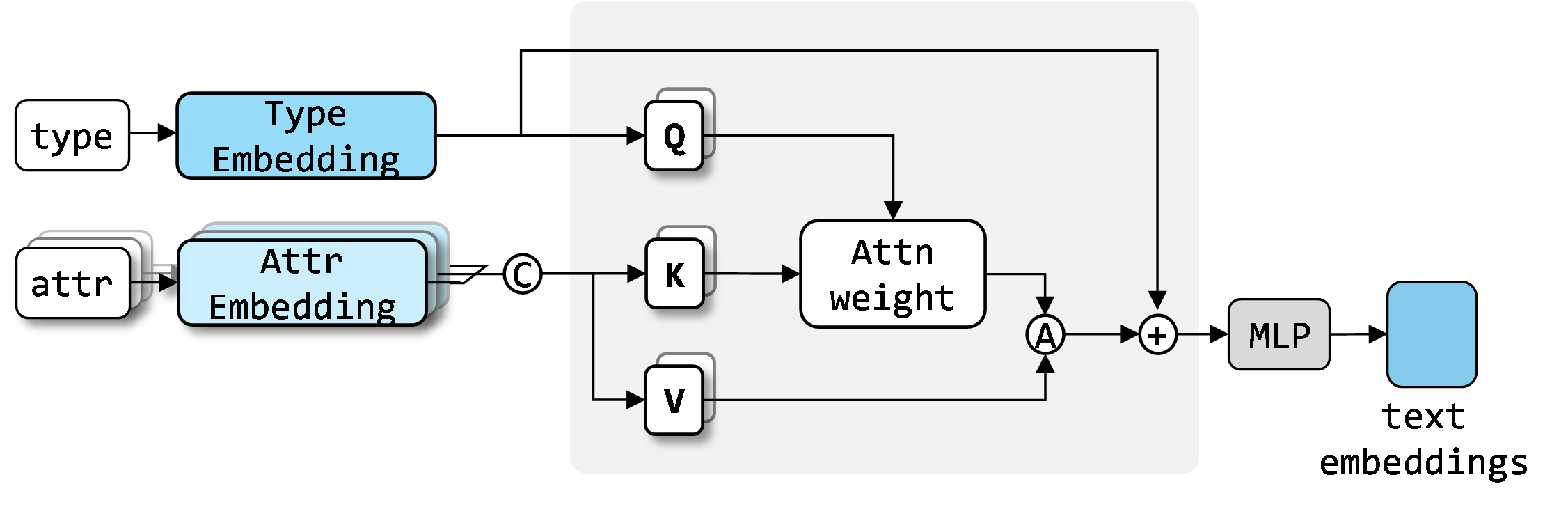}
   \caption{Type-attribute correlation encoder (TACE). It embeds text annotations by modeling semantic correlations between types and attributes by masked multi-head attention.}
   \label{fig3}
\end{figure}

Specifically, as shown in ~\figref{fig3}, for the text annotations $\mathbf{E}_t$, TACE first encodes their decomposed $\mathcal{T} \in \mathbb{R}^{N_t}$ and $\mathcal{A} \in \mathbb{R}^{{N_t} \times a}$ separately, where $N_t$ denotes the number of text annotations. This process can be written as:
\begin{equation}
\label{eq:type_attr_encoder}
\begin{aligned}
    \mathcal{T}_s &= \text{Embed}(\mathcal{T}), \\
    \mathcal{A}_s &= \text{Concat}(\text{MLP}_j(\mathcal{A}^j)), j\in\{1, \dots ,a\},
\end{aligned}
\end{equation}
where $\mathcal{A}^j \in \mathbb{R}^{N_t}$ denotes the $j$-th attribute of $\mathbf{E}_t$. 
$\text{Embed}(\cdot)$ is a neural embedding layer that maps discrete indices to continuous vectors, while $\text{MLP}_j(\cdot)$ consists of two linear layers with RELU activation between them.
This process yields the type embedding $\mathcal{T}_s \in \mathbb{R}^{{N_t} \times 1 \times D}$ and the concatenated attribute embedding $\mathcal{A}_s \in \mathbb{R}^{{N_t} \times a \times D}$, where $D$ denotes the embedding dimension.

Then, $\mathcal{T}_s$ and $\mathcal{A}_s$ are fed into a fusion module to model the latent semantic dependencies between the type and its attributes. 
We adopt a masked multi-head attention, where the type embedding provides the queries and the attribute embedding provides the keys and values. 
Formally, for the $h$-th head ($h=1,\dots,H$), we project
\begin{equation}
\mathbf{Q}_h = \mathcal{T}_s \mathbf{W}^Q_h,\quad
\mathbf{K}_h = \mathcal{A}_s \mathbf{W}^K_h,\quad
\mathbf{V}_h = \mathcal{A}_s \mathbf{W}^V_h,
\end{equation}
and compute the masked attention weights as:
\begin{equation}
\label{eq:multi-head-weight}
\mathbf{W}_h = \mathrm{softmax}\!\left(\frac{\mathbf{Q}_h \mathbf{K}_h^\top}{\sqrt{D}} \odot \mathcal{M}\right),
\end{equation}
where $H$ is the number of heads, and $\mathcal{M} \in \mathbb{R}^{N_t \times 1\times a}$ is the validity mask for $a$ attributes. 
The outputs of all heads are then concatenated and fused with the type embedding to obtain the final semantic embedding $\mathbf{S}_t \in \mathbb{R}^{{N_t}\times 1 \times D}$:
\begin{equation}
\label{eq:final-semantic}
\mathbf{S}_t = \mathrm{MLP}\big( \mathrm{Concat}(\mathbf{W}_1 \mathbf{V}_1, \dots, \mathbf{W}_H \mathbf{V}_H) + \mathcal{T}_s \big).
\end{equation}

Since text annotations also carry corresponding geometric cues, we further inject their geometric features $\mathbf{F}_t$ (e.g., annotation angle and length) into semantic embeddings $\mathbf{S}_t$ to yield initial textual embeddings $\mathbf{X}_t^0 \in \mathbb{R}^{{N_t}\times{D}}$:
\begin{equation}
\label{eq:final-text-embeddings}
\begin{aligned}
    \mathbf{X}_t^0 = \mathbf{S}_t + \text{MLP}(\mathbf{F}_t).
\end{aligned}
\end{equation}
The embedding process of $\mathbf{F}_t$ is consistent with that of $\mathbf{F}_g$, as detailed below, where each text annotation corresponds to a single line.

\subsubsection{Primitives encoder.}
To preserve the geometric continuity of primitives for accurate shape representations, following the prior work~\cite{wei2026point}, we decompose each primitive $e_g^{\,i} \in \mathbf{E}_g$ into a set of lines:
\begin{equation}
\label{eq:line-decomposition}
\begin{aligned}
\mathcal{K}_g^{\,i} = \{\mathit{k}_{\,g}^{i,j}\}_{j=1}^{n_g^{\,i}},
\end{aligned}
\end{equation}
where $n_g^{\,i}$ denotes the number of lines contained in the $i$-th graphical primitive $e_g^{\,i}$. 
For the $j$-th line $\mathit{k}_{\,g}^{i,j}$ in $e_g^i$, we construct its geometric feature as:
\begin{equation}
\label{eq:line-feat}
\begin{aligned}
\mathbf{f}_g^{i,j} = (b^{i,j}, d_{x}^{i,j}, d_y^{i,j}, c_x^{i,j}, c_y^{i,j}, c_x^{\,i}, c_y^{\,i}),
\end{aligned}
\end{equation}
where $b^{i,j}$ denotes the length of the line, $(d_x^{i,j}, d_y^{i,j})$ represents its direction unit vectors, $(c_x^{i,j}, c_y^{i,j})$ is the midpoint of $\mathit{k}_{\,g}^{i,j}$ and $(c_x^{\,i}, c_y^{\,i})$ denotes the geometric centroid of primitive $e_g^{\,i}$. 
Then, $\mathbf{F}_g = \{\mathbf{f}_g^{i,j} \mid e_g^{\,i} \in \mathbf{E}_g,\; \mathit{k}_{\,g}^{i,j} \in \mathcal{K}_g^{\,i}\}$ 
is fed into a primitive encoder (MLP) to produce initial primitive-line embeddings:
\begin{equation}
\mathbf{X}_g^0  = \mathrm{MLP}(\mathbf{F}_g) \in \mathbb{R}^{N_g^0 \times D},
\end{equation}
where $D$ matches textual embedding dimension and $N_g^0=\sum_{i=1}^{|\mathbf{E}_g|} n_g^{\,i}$.

\subsection{Semantic Hierarchy Alignment framework}
Since textual annotations cover multiple semantic levels and correspond to graphical primitives at different geometric scales, naively injecting high-level annotations into low-level primitives blurs and suppresses the intended semantic cues.
We aim to align semantic hierarchies between two modalities for effective cross-modal fusion, allowing annotations to guide the refinement of corresponding primitives.
To this end, we propose the \textit{Semantic Hierarchy Alignment Framework} through a down-upsampling architecture. 

Specifically, to aggregate low-level primitives features into large-scale representations with higher semantics (e.g., progressively merging discrete primitive features into instance-level representations such as doors and further into region-level such as rooms), we adopt Point Transformer V3~\cite{wu2024point} to hierarchically downsample primitive-lines, reducing set cardinality while aggregating local neighborhood features. 
In parallel, a \textit{Multi-level Semantic Filtering} (MSF) mechanism is applied to select textual features that are semantically aligned with each hierarchy level of primitive representations. 
Both the downsampling and filtering processes run in parallel for $L$ layers.
The aligned textual and primitive representations are then fused at corresponding layers to achieve fine-grained semantic interaction and feature enhancement.

\subsubsection{Primitive-line downsampling.}
Hierarchical downsampling aggregates local primitive features into higher-level representations; subsequent upsampling restores the resolution, giving discrete primitives a broader context. At the $l$-th downpooling layer:
\begin{equation}
    \mathbf{X'}_g^{\,l}
= \mathrm{Down}_l(\mathbf{P}_g^{\,l},\,\mathbf{X}_g^{\,l};\, \gamma_l),
\end{equation}
yielding $\mathbf{X'}_g^{\,l}\in\mathbb{R}^{N_g^{l+1}\times D^{l+1}}$. 
Here, $\mathrm{Down}_l(\cdot)$ denotes a downsampling operation implemented via grid-based pooling~\cite{wu2024point} with the grid size rate $\gamma_l$, which partitions primitives into voxel clusters and aggregates features within each voxel to produce a higher-level summary. $N_g^{l+1}$ denotes the number of voxel clusters after downsampling operation, and $\mathbf{P}_g^{\,l}$ denotes the representative positions of clusters and is used to establish neighborhood relations.

\subsubsection{Multi-Level Semantic Filtering.}
As geometric structures are aggregated and the semantic level of primitive features increases through downsampling, textual annotations need to be adaptively filtered to align with corresponding semantic-hierarchy, enabling cross-modal semantic fusion.
To address this, we propose a Multi-level Semantic Filtering (MSF) mechanism. Coupled with the progressive downsampling of primitives, MSF uses primitive features as guidance to hierarchically filter textual features at the corresponding geometric scales, thereby mitigating cross-level interference and preserving essential annotation semantics.

\begin{figure}[tbp]
  \centering
   \includegraphics[width=0.8\linewidth]{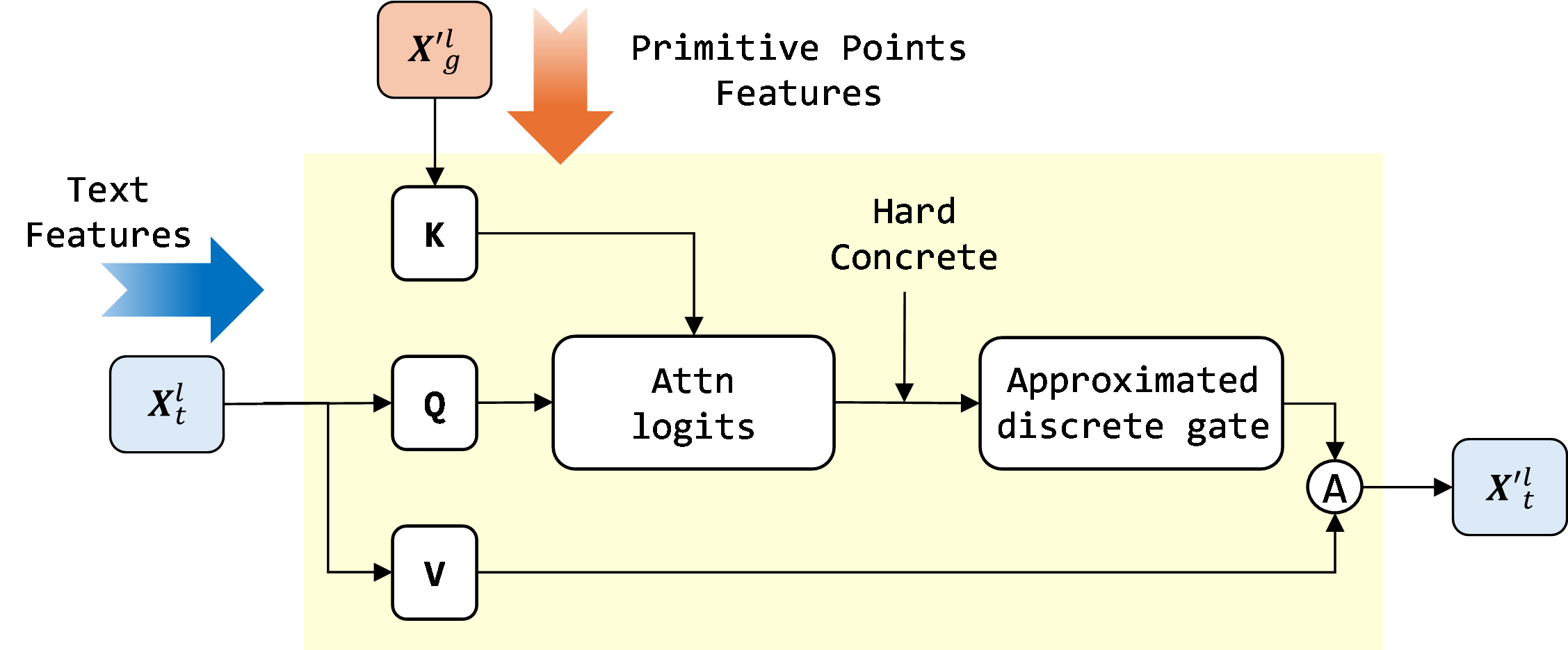}
   \caption{Multi-level semantic filtering layer(MSF). It hierarchically filters textual features guided by primitive representations to achieve semantic alignment across modalities at multiple levels.}
   \label{fig4}
\end{figure}

Specifically, as shown in \figref{fig4}, during the $l$-th semantic filtering layer among the total $L$ layers, the downsampling primitive-line features $\mathbf{X'}_{g}^{\,l}$ are used as guidance to filter the textual features $\mathbf{X}_t^{l} \in \mathbb{R}^{N_t\times D^{l+1}}$ which are projected through $\mathbf{X}_t^{l} = \mathrm{MLP}_l(\mathbf{X}_t^{0})$. The model first computes the cross-modal attention logits $\mathbf{C}^l \in \mathbb{R}^{N_t\times{N_g^{l+1}}}$ between $\mathbf{X}_t^{l}$ and $\mathbf{X'}_{g}^{\,l}$. 
Then, the maximum value over the primitive dimension is taken to obtain the semantic relevance $\mathbf{r}^l \in \mathbb{R}^{N_t}$ of each textual annotation with respect to the primitive lines at the corresponding hierarchy, formulated as:
\begin{equation}
\label{eq:cross-attn}
\begin{aligned}
    \mathbf{C}^l &= \frac{(\mathbf{X}_t^l \mathbf{W}_q) (\mathbf{X'}_{g}^{\,l} \mathbf{W}_k)^T}{\sqrt{D^{l+1}}}, \\
    \mathbf{r}^l_m &= \max_{1 \leq n \leq N_g^{l+1}} C^l_{m,n}, 
    \quad m = 1, \dots, N_t,
\end{aligned}
\end{equation}
where $\mathbf{W}_q$ and $\mathbf{W}_k$ denote projection matrices. $C^l_{m,n} \in \mathbf{C}^l$ is the semantic correlation between $m$-th text annotation and $n$-th primitive-line at layer $l$.

To enable differentiable and near-binary filtering,
we apply hard-concrete relaxation~\cite{louizos2018learning} to semantic relevance scores $\mathbf{r}^l$. The gating-based selection is:
\begin{equation}
\label{eq:hard-concrete}
\begin{aligned}
     \mathbf{G}^l &= \text{HardConcrete}(\mathbf{r}^l, \mathbf{g}), \\
     \mathbf{X'}_t^{\,l} &= \mathbf{G}^l \odot (\mathbf{X}_t^l \mathbf{W}_v).
\end{aligned}
\end{equation}
where $\mathbf{g}\in\mathbb{R}^{N_t}$ is Gumbel noise~\cite{jang2017categorical} injected for stochastic exploration, $\mathbf{G}^l\in[0,1]^{N_t}$ approximates binary gates (near 0: suppressed; near 1: activated), and $\mathbf{W}_v$ is a learnable projection. This formulation preserves end-to-end differentiability while sparsifying the selected textual features and retaining those most relevant to the current hierarchy.
$\mathbf{X'}_t^{\,l}$ denotes the semantically aligned textual features.
Detailed procedures are given in \ref{sec:appendix-msf}.

With the primitive-feature guidance and the hard-concrete relaxation, MSF adaptively aligns text–primitive hierarchies, enabling fusion of more semantically precise textual cues.

\subsubsection{Multimodal fusion.}
After multi-level semantic filtering and downsampling process, taking the $l$-th layer for example, we obtain filtered textual features $\mathbf{X'}_{t}^{\,l}$ which is semantic aligned with primitive-line features $\mathbf{X'}_{g}^{\,l}$. 
We further feed back the semantic aligned $\mathbf{X'}_{t}^{\,l}$ to fuse with $\mathbf{X'}_{g}^{\,l}$ through a cross-attention block:
\begin{equation}
\label{eq:semantic-fusion}
\begin{aligned}
     \mathbf{X}_g^{l+1} = 
     \text{AttnBlock}(
     \text{SerialAttn}(\mathbf{X'}_{g}^{\,l}), \mathbf{X'}_{t}^{\,l}),
\end{aligned}
\end{equation}
where $\text{SerialAttn}(\cdot)$ denotes serialization attention operation~\cite{wu2024point} in Point Transformer V3 which captures structural dependencies among primitive lines. $\text{AttnBlock}(\cdot)$ represents a cross-attention module with spatial embeddings~\cite{vaswani2017attention} that enable the propagation of textual semantics within the corresponding hierarchical level.
A symmetric upsampling pathway \cite{wu2024point} is then employed to progressively recover primitive lines and features by coupling with corresponding downsampling stages. During the $l$-th upsampling:
\begin{equation}
\label{eq:upsample}
\begin{aligned}
     \tilde{\mathbf{X'}}_{g}^{\,l} &= \text{MLP}(\mathbf{X}_g^{L-l}) + \text{Up}(\mathbf{\tilde{X}}_g^{\,l}), \\
     \tilde{\mathbf{X}}_g^{l+1} &= \text{AttnBlock}(\text{SerialAttn}(\tilde{\mathbf{X'}}_{g}^{\,l}), \mathbf{X'}_{t}^{L-l-1}),
\end{aligned}
\end{equation}
where $\text{Up}(\cdot)$ denotes the upsampling operation implemented via partition-based unpooling operation~\cite{wu2024point} that propagates lower-resolution features to higher-resolution based on voxel partitioning.
$\tilde{\mathbf{X'}}_{g}^{\,l}$ represents the enhanced primitive features after upsampling and $\tilde{\mathbf{X}}_g^{\,l}$ is the preceding lower-resolution features to be propagated. 
Detailed procedures are given in \ref{sec:fusion-decoder}.

\subsubsection{Decoder.} 
The final primitive line features $\tilde{\mathbf{X}}_g^{L}$ are passed to the decoder to obtain the final panoptic spotting results:
\begin{equation}
\label{eq:decoder}
\begin{aligned}
    \mathbf{Y}, \mathbf{Z} = \text{Decoder}(\tilde{\mathbf{X}}_g^{L}),
\end{aligned}
\end{equation}
where $\mathbf{Y}$ is the class prediction and $\mathbf{Z}$ is the instance prediction. 

Specifically, the decoder first applies group-wise pooling to aggregate line features within each primitive, resulting in primitive-level representations~\cite{wei2026point}.
These features are then enhanced through intra-layer feature fusion to capture layer-wise context~\cite{liu2024sympoint}.
Finally, a OneFormer3D-based~\cite{kolodiazhnyi2024oneformer3d} head is employed to generate the final predictions.

\subsection{Loss function}
We adopt an overall loss function formulated as follows:
\begin{equation}
\label{eq:overall-loss}
\begin{aligned}
    \mathcal{L} = 
    \lambda_{sem}\mathcal{L}_{sem} +
    \lambda_{cls}\mathcal{L}_{cls} + \lambda_{bce}\mathcal{L}_{bce} + \lambda_{dice}\mathcal{L}_{dice} +
    \lambda_{c}\mathcal{L}_c,
\end{aligned}
\end{equation}
where the cross-entropy loss $\mathcal{L}_{sem}$ is employed for semantic segmentation~\cite{lecun2015deep} and the classification loss $\mathcal{L}_{cls}$ is a multi-class cross-entropy loss~\cite{lecun2015deep} for instance category prediction. The binary cross-entropy~\cite{ronneberger2015u} $\mathcal{L}_{bce}$ and the Dice loss $\mathcal{L}_{dice}$~\cite{milletari2016v} are combined to supervise instance-mask prediction. 

Besides the task-specific loss, we apply a complexity loss $\mathcal{L}_c$~\cite{louizos2018learning} to regularize the semantic filtering process. By penalizing the expectation of active gates, it encourages the model to retain those most semantically relevant text annotations while suppressing redundant ones, leading to more discriminative and semantically aligned filtering. Details are given in \ref{sec:loss}. 

\section{Experiments}
\label{sec:experi}

In this section, extensive experiments are conducted to demonstrate the superiority of TextCAD. We also perform ablation studies and case studies to further validate different parts of our model.

\subsection{Experimental settings}
We outlines the experimental settings, including the dataset, evaluation metrics, baselines and implementation details.

\subsubsection{Dataset.} 
Our experiments are conducted on two real-world publicly available CAD floor plan datasets: FloorPlanCAD-V2~\cite{fan2021floorplancad} and CubiCasa5K~\cite{kalervo2019cubicasa5k}.

\textbf{FloorPlanCAD-V2} is the general dataset proposed for \textit{panoptic symbol spotting}. Compared with its earlier release FloorPlanCAD\allowbreak-V1, this version contains 15,663 CAD drawings spanning a broader range of real-world architectural scenarios, and, importantly, includes abundant textual annotations that are not available in the earlier version. 
To further evaluate the robustness of our model, we additionally conduct experiments on \textbf{CubiCasa5K} which contains 5,000 CAD floorplans with various textual annotations.
We split datasets into training, validation, and test sets with approximate ratios of $\{6:3:1\}$ for FloorPlanCAD-V2 and $\{8:1:1\}$ for CubiCasa5K. More details on data preprocessing provided in \ref{sec:datasets}.

\subsubsection{Evaluation metrics.}
We evaluate with multiple metrics following prior work~\cite{fan2021floorplancad,fan2022cadtransformer,zheng2022gat,liu2024symbol,wang2025panoramic,liu2026text,wei2026point,luo2026archcad}, including \textbf{PQ} (Panoptic Quality), tailored to panoptic symbol spotting and jointly assessing semantic classification and instance segmentation, as well as \textbf{PQ-Thing} and \textbf{PQ-Stuff} to measure performance on countable \textit{thing} categories and uncountable \textit{stuff} categories, respectively.
Additionally, the standard classification metrics \textbf{F1} and \textbf{wF1} (length-weighted F1) are for semantic spotting. 
Detailed definitions are in the \ref{sec:metrics}.

\subsubsection{Baselines.} 
We compare TextCAD with baselines of multiple paradigms: image-based method PanCADNet~\cite{fan2021floorplancad}, graph-based method GAT-CADNet~\cite{zheng2022gat} and CADTransformer~\cite{fan2022cadtransformer}, point cloud-based method SymPoint~\cite{liu2024symbol}, SymPointV2 \cite{liu2024sympoint} and DPSS~\cite{luo2026archcad}, line-based method VecFormer~\cite{wei2026point} and text-incoporated methods PFL-Net~\cite{wang2025panoramic}, TNet~\cite{liu2026text} and TriNet~\cite{xing2026multimodal}. Additional descriptions are provided in \ref{sec:baselines}. All baselines were reimplemented on both FloorPlanCAD-V2 with textual annotations and CubiCasa5K.

\subsubsection{Implementation details.}
Hyperparameters in TextCAD are extensively searched and set to their optimal values, with parameter sensitivity analysis provided in \ref{sec:param-analysis}.
For the TACE, the number of attributes $a$ is set to 4, the embedding dimension $D$ is set to 32, and the number of multi-heads $H$ is set to 4. 
Within the Semantic Hierarchy Alignment framework, 
both the number of downpooling layers and the filtering layers $L$ are set to 5, with grid size rates $[1, 2, 2, 2, 2]$ for primitives lines.
Notably, the first layer performs no downpooling or semantic filtering; it only performs projection to aligned feature dimensions.

AdamW optimizer with an initial learning rate of 0.0001 and warm-up ratio of 0.05 is employed and training spans 700 epochs. 
All experiments, including baselines, are run three times and the average is reported; training uses four NVIDIA RTX A6000 GPUs with a batch size of 2.

\subsection{Main performance}
Herein, we analyze performances of various baselines and TextCAD under two experimental settings, with the corresponding results reported in Table 1.
``w/ prior'' denotes that the model is provided with prior primitive attributes, namely auxiliary properties of graphical primitives, such as layer assignments or color information, that may provide category-indicative cues.
``w/o prior'' denotes that such prior attributes are not used.

Some methods (e.g. CADTransformer, SymPoint) are not included under the ``w/ prior'' setting because they do not inherently support the use of prior primitive attributes.
Notably, for CubiCasa5K, results under ``w/ prior'' setting are not reported, as the dataset itself lacks such prior information.
For clarity, all results are multiplied by 100.

\begin{table}[tbp]
\centering
\caption{Main Performance of TextCAD Compared with Baselines.}
\label{main-performance}

\scriptsize
\setlength{\tabcolsep}{3.2pt}
\renewcommand{\arraystretch}{1.12}

\resizebox{\linewidth}{!}{%
\begin{tabular}{@{}clccccc@{\hspace{5pt}}ccccc@{}}
\toprule
\multirow{2}{*}{\textit{Setting}}
& \multirow{2}{*}{Method}
& \multicolumn{5}{c}{\textbf{FloorPlanCAD-V2}}
& \multicolumn{5}{c}{\textbf{CubiCasa5K}} \\
\cmidrule(lr){3-7}
\cmidrule(lr){8-12}

&
& PQ
& PQ-Thing
& PQ-Stuff
& F1
& wF1
& PQ
& PQ-Thing
& PQ-Stuff
& F1
& wF1 \\
\midrule

\multirow{11}{*}{w/o prior}
& PanCADNet
& 57.60 & 65.78 & 53.34 & 78.7 & 78.0
& 60.80 & 67.32 & 46.89 & 80.5 & 76.4 \\

& GAT-CADNet
& 71.31 & 73.29 & 58.08 & 84.4 & 81.4
& 78.52 & 79.39 & 55.92 & 89.4 & 84.7 \\

& CADTransformer
& 71.75 & 73.65 & 58.39 & 80.4 & 78.5
& 74.74 & 74.89 & 53.98 & 82.0 & 84.1 \\

& SymPoint
& 83.27 & 86.68 & 57.26 & 85.7 & 84.7
& 89.16 & 90.58 & 50.47 & 93.6 & 88.7 \\

& SymPointV2
& 82.86 & 86.24 & 57.07 & 86.1 & 85.7
& 91.47 & 92.76 & 57.25 & 93.9 & 91.9 \\

& DPSS
& 84.46 & 87.56 & 61.45 & 91.5 & 91.0
& 91.33 & 92.56 & 59.15 & 94.1 & 93.5 \\

& VecFormer
& 88.14 & 87.26 & 89.17 & 87.8 & 89.8
& 94.58 & 95.59 & 86.95 & 96.2 & 96.4 \\

& PFL-Net
& 73.39 & 75.36 & 59.72 & 80.6 & 78.9
& 77.95 & 78.24 & 54.69 & 85.8 & 83.5 \\

& TNet
& 73.85 & 75.93 & 59.44 & 81.0 & 78.9
& 78.54 & 78.75 & 57.21 & 84.6 & 86.5 \\

& TriNet
& 83.98 & 87.25 & 59.23 & 91.2 & 90.2
& 90.31 & 91.66 & 53.69 & 94.1 & 93.4 \\

\cmidrule(lr){2-12}
& \textbf{TextCAD}
& \textbf{91.09}
& \textbf{91.22}
& \textbf{90.97}
& \textbf{91.7}
& \textbf{91.4}
& \textbf{96.53}
& \textbf{97.23}
& \textbf{91.48}
& \textbf{98.1}
& \textbf{98.3} \\

\midrule

\multirow{4}{*}{w/ prior}




& SymPointV2
& 89.34 & 90.55 & 81.01 & 89.3 & 88.7
& -- & -- & -- & -- & -- \\

& DPSS
& 89.39 & 90.54 & 81.31 & 92.2 & 91.5
& -- & -- & -- & -- & -- \\

& VecFormer
& 90.76 & 90.64 & 90.90 & 90.0 & 91.3
& -- & -- & -- & -- & -- \\




\cmidrule(lr){2-12}
& \textbf{TextCAD}
& \textbf{92.67}
& \textbf{93.05}
& \textbf{92.32}
& \textbf{93.4}
& \textbf{91.9}
& -- & -- & -- & -- & -- \\

\bottomrule
\end{tabular}%
}
\end{table}

\subsubsection{TextCAD performance}

Based on ~\tabref{main-performance}, we draw the following key observations:

i) Under both settings, TextCAD consistently achieves the best performance across all metrics, outperforming the strongest baseline by nearly 2 percentage points on overall PQ which serves as the comprehensive evaluation of \textit{panoptic symbol spotting}. This demonstrates TextCAD is able to leverage critical cues within textual annotations effectively to improve overall performances as well as both countable \textit{thing} and uncountable \textit{stuff} categories.

ii) Compared with methods that rely solely on graphical primitives ~\cite{fan2021floorplancad,zheng2022gat,fan2022cadtransformer,liu2024symbol,liu2024sympoint,luo2026archcad,wei2026point}, TextCAD delivers consistent gains across all metrics. 
These improvements stem from leveraging key semantic cues in abundant CAD textual annotations, such as functional descriptions and dimensional specifications, which enrich primitive representations, 
help distinguish incomplete geometry structures or overlapping primitives, and ultimately enable more accurate detection.
In contrast, primitive-only approaches operate from a single, limited modality and overlook these crucial annotation semantics, thereby hindering spotting performance.

iii) Compared with methods that incorporate textual annotations~\cite{wang2025panoramic,liu2026text,xing2026multimodal}, the superiority of TextCAD mainly stems from two aspects: 
(1)~\textit{Accurate and lightweight textual semantic modeling}: instead of relying on language models (e.g., PFL-Net), the proposed TACE adopts a lightweight architecture that explicitly captures type--attribute correlations, thereby producing more robust textual semantic. 
(2)~\textit{Hierarchy-aligned cross-modal fusion}: rather than simply injecting textual annotations via direct concatenation fusion (e.g., TriNet or TNet), TextCAD treats textual annotations as an independent modality and employs MSF to adaptively filter and align cross-modal semantics, enabling multimodal fusion under semantically aligned hierarchies.
Together, these two advantages allow TextCAD to provide more accurate and direct semantic cues.

iv) Under both ``w/o prior'' and ``w/ prior'' settings, TextCAD outperforms existing methods across all metrics, demonstrating strong robustness to prior information variations. 
Particularly on FloorPlan-V2, TextCAD suffers the smallest decrease in overall PQ when prior is unavailable.
Since CubiCasa5K does not provide prior information, its ``w/o prior'' evaluation is not applicable.
This verifies TextCAD can achieve robust predictions by constructing structured textual representations and effectively fusing semantic cues with geometric features, compensating for the absence of prior information.

\begin{figure}[tbp]
  \centering
   \includegraphics[width=0.8\linewidth]{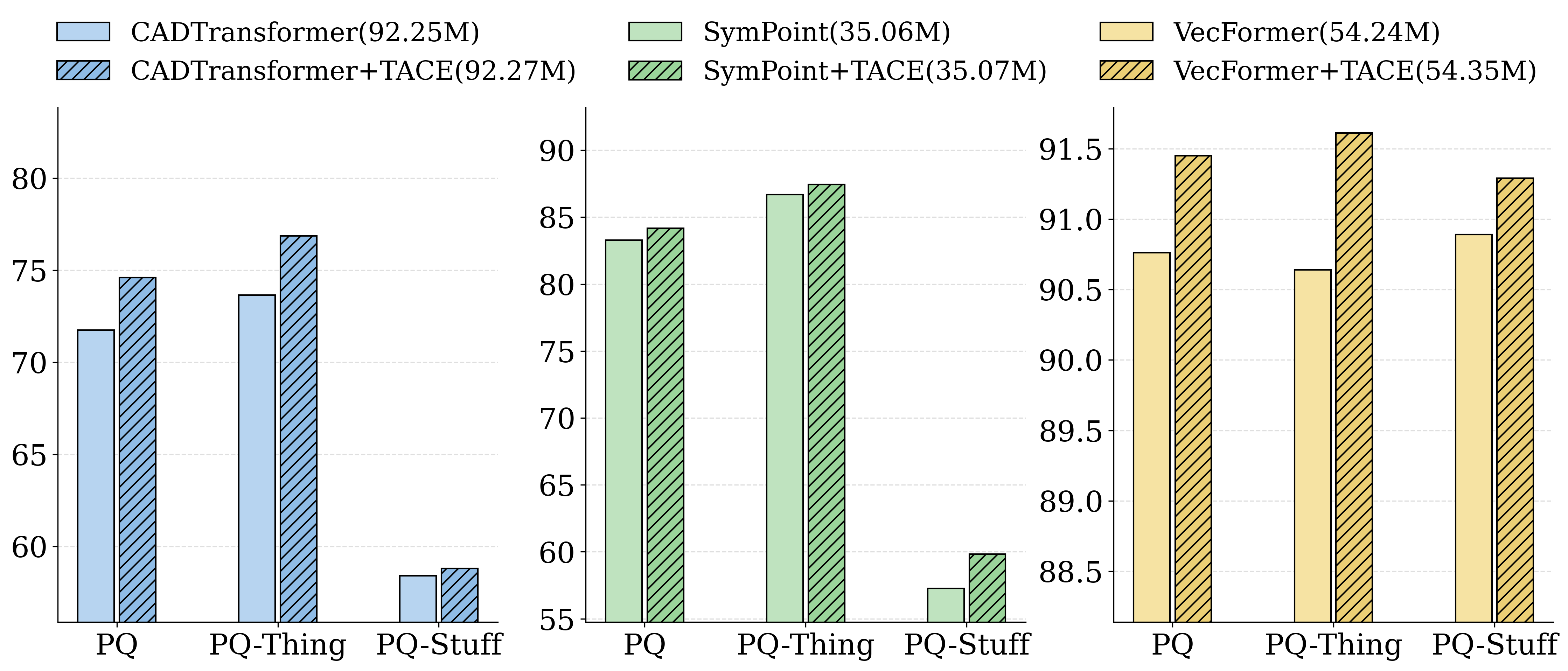}
   \caption{Performance of TACE integrated into different baselines. TACE improves performance across baselines, including CADTransformer, SymPoint, and VecFormer, with negligible additional parameters shown in the legend.}
   \label{fig6}
\end{figure}

\subsubsection{Efficacy analysis}
To assess the impact of TACE, we further integrate it into three representative baselines: CADTransformer~\cite{fan2022cadtransformer}, SymPoint~\cite{liu2024symbol} and VecFormer~\cite{wei2026point}. As shown in ~\figref{fig6}, on FloorPlanCAD-V2, adding TACE yields notable gains without materially increasing parameters, highlighting the value of incorporating annotations and demonstrating TACE’s ability to link type-attribute pairs and embed key semantic cues in a lightweight manner.

\subsection{Ablation study}
We conduct several ablation studies on FloorPlanCAD-V2 to further validate the effectiveness of certain designs in TextCAD.

\subsubsection{Type-Attribute Correlation Encoder module.}
To further verify the necessity of TACE in jointly modeling the semantic correlations between type and attributes, we replace TACE by ``TypeEmb'' (encoding only the type via embedding layer), ``TypeAttrMLP'' (concatenating type and attributes followed by MLP) and ``BertEmb'' (encoding textual content using BERT\cite{devlin2019bert}). As shown in ~\tabref{tab:TACE-ablation2}, TACE consistently achieves the best performance, confirming the necessity of jointly capturing latent semantic correlations between type and attributes. In addition, TACE remains lightweight instead of employing the tokenizer of LM.

\begin{table}[t]
\centering
\caption{Comparison of TACE with Different Encoding Methods.}
\label{tab:TACE-ablation2}
\renewcommand{\arraystretch}{1.1}
\resizebox{0.8\columnwidth}{!}{
\begin{tabular}{@{}lcccccc@{}}
\toprule
Method
& PQ
& PQ-Thing
& PQ-Stuff
& F1
& wF1
& Params \\
\midrule
TypeEmb
& 91.25 & 91.07 & 91.43 & 90.5 & 91.2 & 54.34M \\

TypeAttrMLP
& 91.16 & 91.18 & 91.13 & 90.6 & 91.1 & 54.47M \\

BertEmb
& 91.13 & 91.24 & 91.01 & 90.9 & 91.4 & 156.65M \\

\textbf{TACE}
& \textbf{91.69}
& \textbf{91.84}
& \textbf{91.54}
& \textbf{91.7}
& \textbf{91.6}
& 54.35M \\
\bottomrule
\end{tabular}
}
\end{table}

\subsubsection{Multi-Level Semantic Filtering mechanism.}
To evaluate the effectiveness of semantic hierarchy alignment, we ablate the MSF design in ~\tabref{tab:MSF-ablation}.
Notation: `Text–GP' treats textual annotations as part of the graphical-primitive modality; `Text–Uni.' feeds text as a separate unimodal input; `Multi-L.' enables text injection at multiple hierarchy levels; `Filt.' applies multi-level semantic filtering; `GP-G.' uses primitive features to guide filtering.

\begin{table}[htbp]
\centering
\caption{Ablation Study of Semantic Filtering.}
\label{tab:MSF-ablation}
\resizebox{\linewidth}{!}{%
\begin{tabular}{cccccccccc}
\toprule
Text-GP & Text-Uni. & Multi-L. & Filt. & GP-G. &
PQ & PQ-Thing & PQ-Stuff & F1 & wF1 \\ 
\midrule
\checkmark &  &  &  &  & 91.69 & 91.84 & 91.54 & 91.7 & 91.6 \\
 & \checkmark &  &  &  & 91.77 & 91.81 & 91.73 & 91.2 & 91.7 \\
 & \checkmark & \checkmark &  &  & 91.50 & 90.83 & 92.08 & 90.9 & 91.4 \\
 & \checkmark & \checkmark & \checkmark &  & 92.12 & 92.27 & 91.98 & 91.9 & 91.7 \\
 & \checkmark & \checkmark & \checkmark & \checkmark & \textbf{92.67} & \textbf{93.05} & \textbf{92.32} & \textbf{93.4} & \textbf{91.9} \\
\bottomrule
\end{tabular}%
}
\end{table}

From ~\tabref{tab:MSF-ablation}, we derive the following key observations: treating CAD drawings as multi-modal (rather than folding text into the GP stream) yields better performance; however, naively injecting text at multiple levels without filtering degrades accuracy because misaligned text introduces cross-level noise. 
Enabling multi-level semantic filtering mitigates this issue and improves spotting performance, and guiding the filtering with primitive features (GP-G.) provides the best alignment and further gains by directly coupling textual semantics with primitive geometry.

More validation of semantic hierarchy alignment is provided in \ref{sec:msf-case-study}.

\subsection{Case study}
\begin{figure}[htbp]
    \centering

    \begin{subfigure}[b]{0.185\linewidth}
        \centering
        \includegraphics[width=\linewidth]{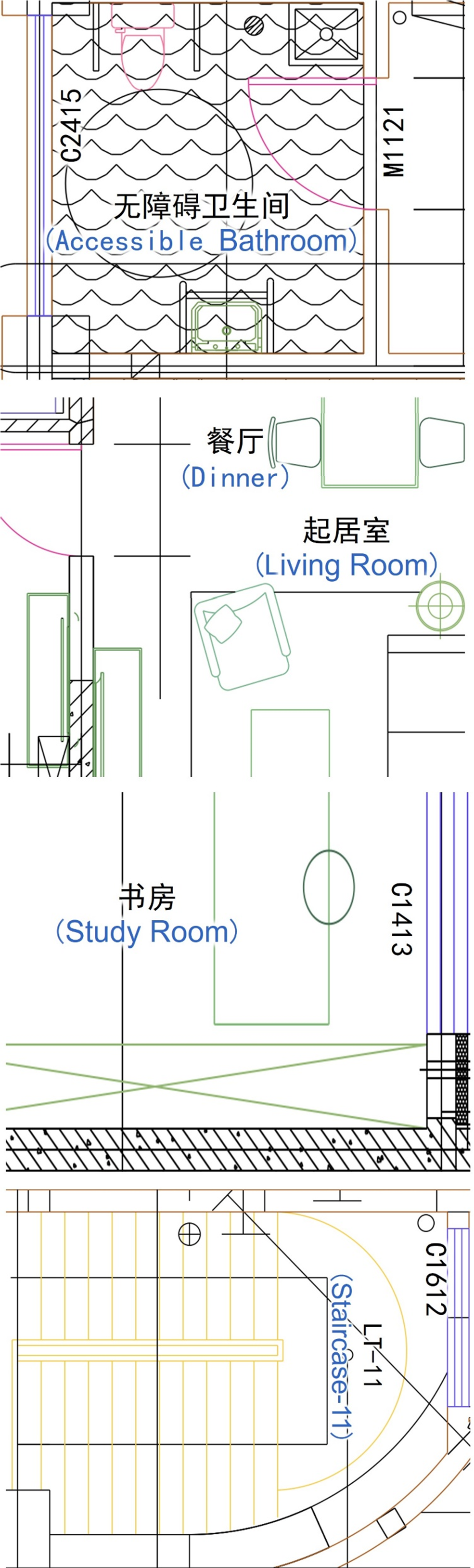}
        \caption{GT}
        \label{sub-fig:GT}
    \end{subfigure}%
    \hfill
    \vrule width 0.1pt
    \hfill
    \begin{subfigure}[b]{0.185\linewidth}
        \centering
        \includegraphics[width=\linewidth]{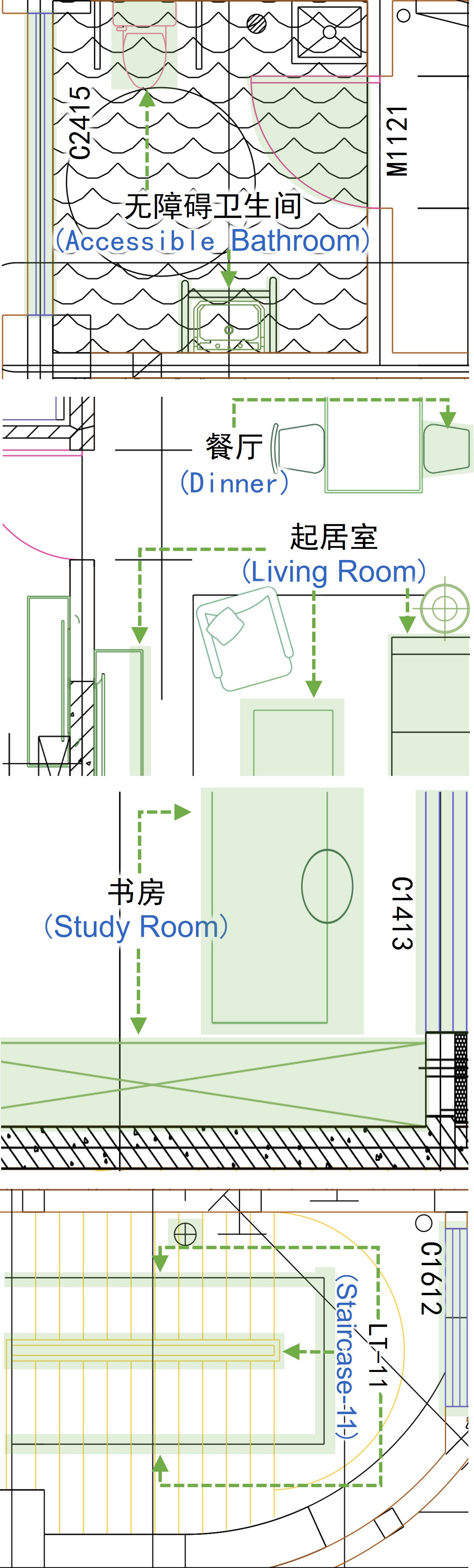}
        \caption{TextCAD}
        \label{sub-fig:TextCAD}
    \end{subfigure}%
    \hfill
    \vrule width 0.1pt
    \hfill
    \begin{subfigure}[b]{0.185\linewidth}
        \centering
        \includegraphics[width=\linewidth]{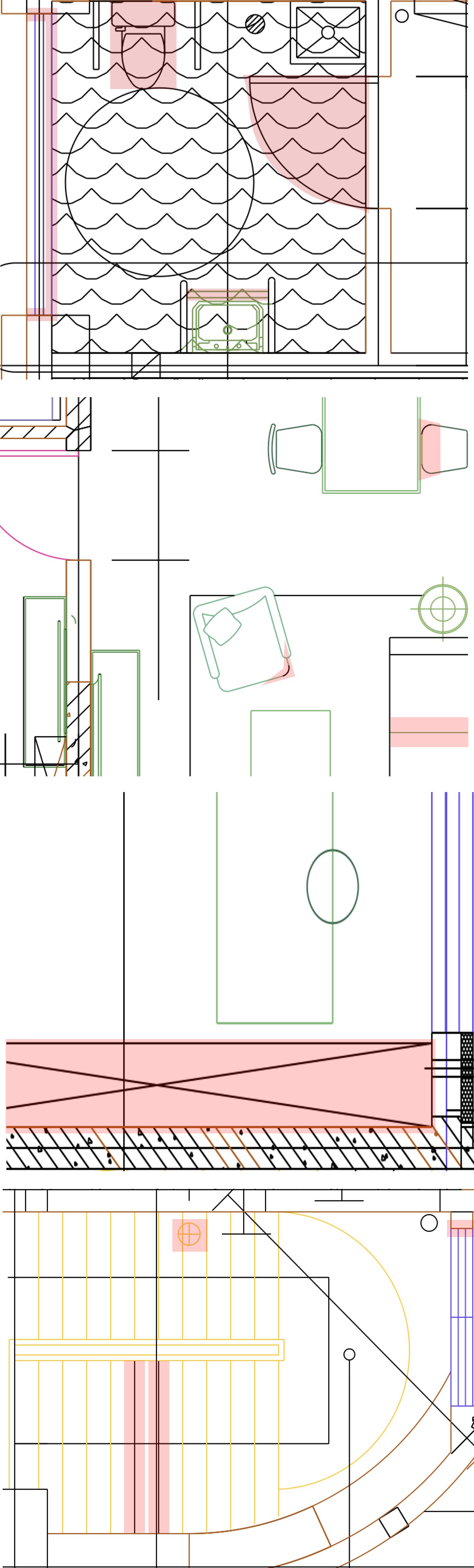}
        \caption{CADTrans.}
        \label{sub-fig:CADTransformer}
    \end{subfigure}%
    \hfill
    \vrule width 0.1pt
    \hfill
    \begin{subfigure}[b]{0.185\linewidth}
        \centering
        \includegraphics[width=\linewidth]{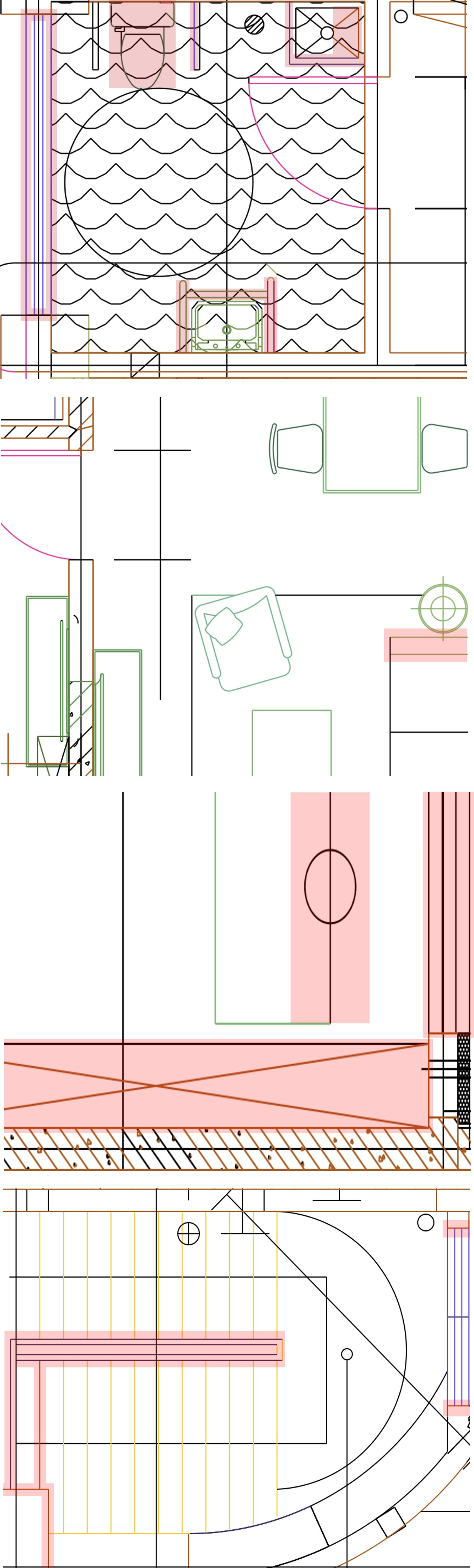}
        \caption{SymPoint}
        \label{sub-fig:SymPoint}
    \end{subfigure}%
    \hfill
    \vrule width 0.1pt
    \hfill
    \begin{subfigure}[b]{0.185\linewidth}
        \centering
        \includegraphics[width=\linewidth]{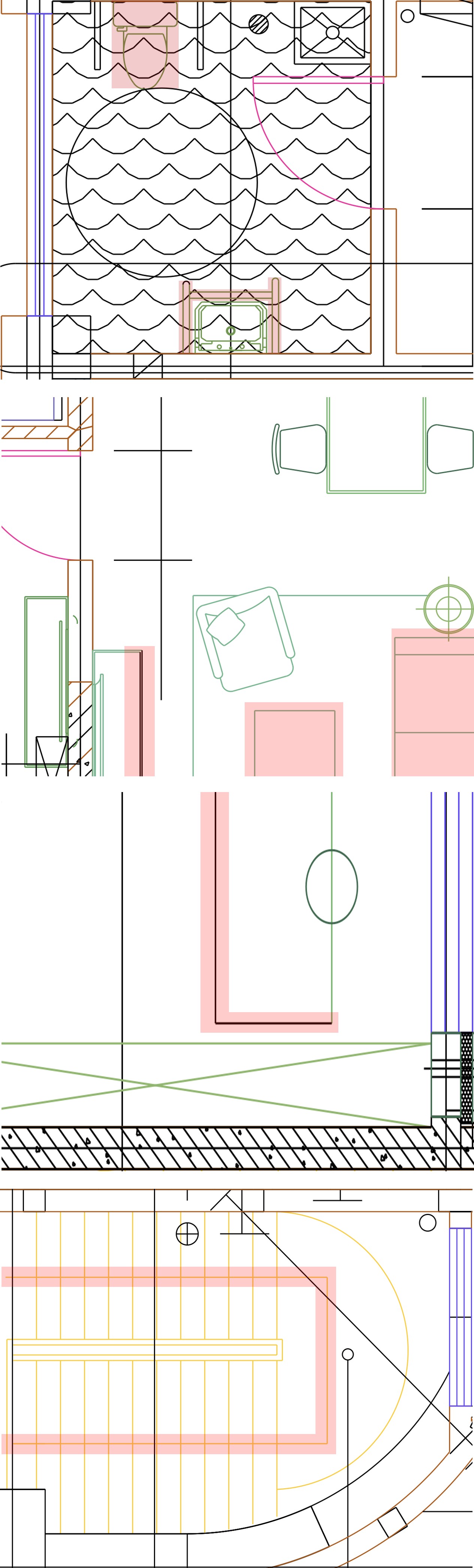}
        \caption{VecFormer}
        \label{sub-fig:VecFormer}
    \end{subfigure}

    \caption{Qualitative comparison of different methods. TextCAD produces more accurate predictions, particularly in regions with partially incomplete or densely overlapping structures. Its results are better
    aligned with the ground truth than other models.}
    \label{fig:qualitative-comparison}
\end{figure}

The qualitative results on FloorplanCAD-V2 are visualized in ~\figref{fig:qualitative-comparison}, which includes (a) ground truth, (b) predictions from TextCAD, and (c/d/e) those from three representative baselines.
In~\figref{sub-fig:TextCAD}, green regions highlight key regions of interest, and arrows represent results within these regions are affected by corresponding text annotations. Red regions in ~\figref{sub-fig:SymPoint}, ~\figref{sub-fig:CADTransformer} and ~\figref{sub-fig:VecFormer} indicate wrong predictions from baselines models.

It can be clearly observed that TextCAD shows a high degree of consistency with the ground truth (GT).
In highlighted regions, TextCAD demonstrates a clear advantage over baselines by explicitly leveraging the semantic cues contained in annotations. 
Specifically, TextCAD achieves more robust and accurate results, particularly in complex regions with dense or overlapping line structures, or instance regions where graphical structures are partially incomplete. 
For instance, `\texttt{Washroom}' helps accurately identify toilet-related primitives through interaction with corresponding graphical elements, while `\texttt{Staircase-11}' enables precise recognition of structures such as stairs and handrails by capturing the underlying semantic relationships indicated by the annotations.
Additional results are provided in \ref{sec:qualitative-result}.

\section{Related work}
\label{sec:related-work}

Various research tasks were explored on CAD drawings~\cite{fan2021floorplancad,yang2023vectorfloorseg,xing2025comprehensive,yan2025substation}. We focus on \textit{panoptic symbol spotting}
which was first proposed in~\cite{fan2021floorplancad} to address the limitation of focusing solely on countable things while neglecting uncountable stuffs~\cite{rezvanifar2019symbol}. Existing methods fall into the following categories:

Early studies take conventional computer vision methods~\cite{fan2021floorplancad,pang2024pixel,rezvanifar2020symbol} by converting CAD drawings into raster images. PanCADNet~\cite{fan2021floorplancad} employs Faster R-CNN for instance recognition and Graph Convolutional Network (GCN) for semantic detection. However, converting vector primitives to pixels losses geometric precision and fine-grained structural details.

Subsequent studies decompose vector drawings into primitives and leverage Transformer-based~\cite{fan2022cadtransformer} or GNN-based methods~\cite{zheng2022gat,zhao2025architecad,carrara2025vectorgraphnet}. CADTransformer~\cite{fan2022cadtransformer} modifies Vision Transformer~\cite{dosovitskiy2020image} to update primitive features,
while GAT-CADNet~\cite{zheng2022gat} models primitives as graph nodes by employing Graph Attention Network (GAT) for feature propagation. Vector-based methods better preserve internal CAD structures but face challenges on complex drawings with numerous primitives due to memory constraints.

SymPoint~\cite{liu2024symbol} treats CAD drawings as point cloud by abstracting primitives as point sets to enhance feature extraction. Subsequent studies follow this direction  ~\cite{liu2024sympoint,yang2024cadspotting,luo2026archcad}, but emphasize only on graphical features and structural relations, neglecting the semantics in textual annotations.

Recently, VecFormer~\cite{wei2026point} proposed line-based representation to preserves the geometric continuity and enhance shape representation, but still stranded by the same issues that textual annotations semantics had been ignored.

A few studies incorporate textual annotations but via language models such as BERT tokenizer~\cite{wang2025panoramic} and CLIP text encoder~\cite{wang2025enhanced}. Others either naively concatenate text categories to nearby primitive features~\cite{xing2026multimodal} or simply embeds them from feature maps without structured semantic modeling or effective cross-modal fusion~\cite{liu2026text}. They overlook the hierarchical semantics and syntactic structure of annotations, limiting their semantic.

In contrast, our method syntactically represents textual annotations to capture latent semantics, and performs multi-level semantic filtering for cross-modal alignment and fusion, thereby enabling semantically enriched for discriminative primitive representations.

\section{Conclusion}
\label{sec:conclusion}

In this work, we propose \textbf{TextCAD}, a multi-modal framework that jointly integrates textual annotations and graphical primitives for \textit{panoptic symbol spotting}. The proposed Type--Attribute Correlation Encoder (TACE) embeds rich annotation semantics by explicitly modeling intrinsic correlations between the type and attributes, yielding expressive semantic representations. The Semantic Hierarchy Alignment framework applies Multi-level Semantic Filtering (MSF) with primitive downsampling, enabling semantically hierarchy-consistent \allowbreak cross-modal alignment for effective modality fusion.
Experiments show that \allowbreak TextCAD surpasses prior state of the art by nearly 2 percentage points on standard metrics, validating the benefit of textual annotations incorporation. Ablation studies and case analyses further confirm the effectiveness and lightweight nature of TACE, and the necessity of MSF.
Moreover, TextCAD generalizes well to diverse annotation styles: TACE accommodates varied type-attribute patterns, and the alignment module adapts across semantic levels. 
Future work will focus on more effective cross-modal fusion to improve reliability.

\clearpage
\appendix
\section{Further discussion}
\label{sec:further discussion}

Here we discuss the type-attribute decomposing and the extensibility.

\subsection{Type-attribute decomposing} 
\label{sec:tace-decompose}

We perform preprocessing on textual annotations to extract their type and attribute information. 

Based on the observation and syntactic analysis of CAD textual annotations, we categorize them into various \emph{types} according to their syntactic structure and associated attribute patterns. Each type is paired with a specific set of attributes, forming distinct \textit{type–attribute patterns}. For instance, type \texttt{Elevator} is attached by a numeric attribute denotes load capacity, and type \texttt{Door} is paired with two numeric attributes represent dimensions. The number of attributes $a$ is set to 4, while among four attributes $\mathcal{A}^j, j\in\{1\dots4\}$, $\mathcal{A}^1$, $\mathcal{A}^2$ and $\mathcal{A}^3$ denote numerical information and $\mathcal{A}^4$ is grade information.

We design an automated parsing framework to separate CAD text annotations into their corresponding types $\mathcal{T}$ and attributes $\mathcal{A}^j$. Based on predefined type–attribute structures, our tool employs regular-expression~\cite{friedl2006mastering} based pattern matching to automatically identify and extract the attributes associated with different type.
This process yields 166 types of textual annotation with specific attribute combinations for FloorPlanCAD-V2 and 86 types for CubiCasa5K. We present 13 representative examples in~\tabref{tab:type-attr}.

\begin{table}[htbp]
\centering
\caption{Examples of type--attribute patterns.}
\label{tab:type-attr}
\renewcommand{\arraystretch}{1.1}
\setlength{\tabcolsep}{3pt}
\resizebox{\textwidth}{!}{
\begin{tabular}{@{}c*{16}{c}@{}}
\toprule
\textbf{Attribute}
& \texttt{Number} & \texttt{NumberDeci} & \texttt{Multi} & \texttt{Door} & \texttt{FMDoor} & \texttt{Elevator} & \texttt{Window} & \texttt{FMWindow} & \texttt{Slope} & \texttt{Weight} & \texttt{Floor} & \texttt{KW} & \texttt{Kitchen} \\
\midrule

\textbf{Attr1}
& Integer & Decimal & length & width & width & load & width & width & gradient & kilogram & floor & kilowatt & -- \\

\textbf{Attr2}
& -- & -- & width & height & height & -- & height & height & -- & -- & -- & -- & -- \\

\textbf{Attr3}
& -- & -- & height & -- & -- & -- & -- & -- & -- & -- & -- & -- & -- \\

\textbf{Attr4}
& -- & -- & -- & -- & grade & -- & -- & grade & -- & -- & -- & -- & -- \\

\bottomrule
\end{tabular}
}
\end{table}

The extracted type $\mathcal{T}$ and attributes $\mathcal{A}^j$ are independently embedded. The numerical attributes $\mathcal{A}^1$, $\mathcal{A}^2$ and $\mathcal{A}^3$ are embedded using the corresponding $\text{MLP}_j$, while grade attribute $\mathcal{A}^4$ is encoded via an neural embedding layer.

The proposed type-attribute representation schema can be explained by the concept of factorized representation: complex semantics are typically composed of multiple underlying factors of variation~\cite{bengio2013representation}, and explicitly disentangling these factors improves both expressiveness and generalization.

Our study formulates CAD textual annotations as type–attribute syntactic structure. This formulation can be viewed as a factorized representation, which decomposes entangled semantics into multiple composable subspaces, thereby reducing representation complexity and improving generalization.



\subsection{Extensibility and robustness} 
\label{sec:tace-extensibility}

TACE and the unified type–attribute representation schema are not tied to a specific dataset; instead, it exhibits strong generality and extensibility.

Specifically, the defined types capture representative textual annotation patterns in CAD floor plans. Rather than a fixed taxonomy, this type--attribute scheme provides a general structured representation framework that can be readily extended to other datasets by adding new types and defining their corresponding attributes structures under the principles of this schema.

Additionally, in real-world design, CAD annotations typically follow engineering drafting conventions, resulting in regular and semi-structured patterns. TACE and its pipeline are specifically designed for such realistic and regular patterns. The robustness of this design is further supported by experimental results on two real-world CAD datasets.
Moreover, unmatched cases during TACE decomposition are mostly due to anomalous human annotations, which are filtered to avoid noisy semantics. This information loss is also limited, as neighboring textual cues provide complementary context.

\section{Model details}
\label{sec:model-detail}

We make detailed supplements to our model, including multi-level semantic filtering, multimodal fusion, decoder, and loss function.

\subsection{Multi-Level Semantic Filtering} 
\label{sec:appendix-msf}
Semantic filtering employs hard-concrete relaxation to obtain approximated binary gates through the differentiable process. 
During the $l$-th semantic filtering layer ($l=0\cdots L-1$), we first introduce randomness into $\mathbf{r}^l$ by adding Gumbel noise $\mathbf{g}$~\cite{jang2017categorical} for stochastic exploration, and then obtain a smooth activation probability $\mathbf{s}^l \in \mathbb{R}^{N_t}$, formulated as:
\begin{equation}
\label{eq:gumbel-noise}
\begin{aligned}
    \mathbf{g} &= -\log(-\log(\mathbf{u})), \mathbf{u} \sim \mathrm{Uniform}(0,1)^{N_t}\\
    \mathbf{s}^l &= \sigma\Big(\frac{\mathbf{r}^l + \mathbf{g}}{\tau}\Big),
\end{aligned}
\end{equation}
where $\tau$ and $\sigma$ denote the temperature and the \text{softmax}, respectively. The discrete gating behavior is then further approximated as:
\begin{equation}
\label{eq:stretch-clamp}
\begin{aligned}
    \mathbf{G}^l  = \text{min}(1, \text{max}(0, \mathbf{s}^l (\zeta - \gamma) + \gamma)),
\end{aligned}
\end{equation}
where $\gamma$ and $\zeta$ are the lower and upper predefined stretch limits. The approximated binary gates $\mathbf{G}^l \in [0,1]^{N_t}$ is generated by first linearly mapping $\mathbf{s}^l$ to an extended interval $(\gamma, \zeta)$ and then clipping it to the range of $[0,1]$.

\subsection{Multimodal fusion and decoder} 
\label{sec:fusion-decoder}
Here we make a detailed supplement to the process of multimodal fusion, primitive-line upsampling and the decoder.

\subsubsection{Multimodal fusion}
\label{appendix:multimodal-fusion}
After the $l$-th multi-level semantic filtering and downsampling process, we obtain filtered textual features $\mathbf{X'}_{t}^{\,l}$ which is semantic aligned with primitive-line features $\mathbf{X'}_{g}^{\,l}$. 
we fuse $\mathbf{X'}_t^{\,l}$ back with $\mathbf{X'}_g^{\,l}$:
\begin{equation}
\label{eq:app-semantic-fusion}
\begin{aligned}
     \mathbf{X}_g^{l+1} = 
     \text{AttnBlock}(
     \text{SerialAttn}(\mathbf{X'}_{g}^{\,l}), \mathbf{X'}_{t}^{\,l}),
\end{aligned}
\end{equation}
where $\text{SerialAttn}(\cdot)$ denotes the serialization attention operation in Point Transformer V3 which performs self-attention within patches derived from the serialized point sequence~\cite{wu2024point}.

$\text{AttnBlock}(\cdot)$ is implemented as a cross-modal attention module, where primitive-line features $\mathbf{X'}_g^{\,l}$ serve as queries and textual features $\mathbf{X'}_t^{\,l}$ serve as keys and values. This enables textual semantics and geometric representations to interact at the same hierarchical level.
To enhance spatial awareness, spatial embeddings are further incorporated into attention computation and feature aggregation, allowing nearby textual features to exert stronger influence on each primitive and promoting more semantically relevant 
cross-modal interactions.

\subsubsection{Primitive-line upsampling}
\label{sec:appendix-upsampling}

The upsampling structure is symmetric to downsampling stages. At the $l$-th upsampling layer, the low-resolution features $\tilde{\mathbf{X}}_g^{\,l}$ are interpolated onto higher-resolution:
\begin{equation}
\label{eq:interplote}
\begin{aligned}
     \tilde{\mathbf{X'}}_{g}^{\,l} &= \text{MLP}(\mathbf{X}_g^{L-l}) + \text{Up}(\mathbf{\tilde{X}}_g^{\,l}).
\end{aligned}
\end{equation}
$\text{Up}(\cdot)$ denotes an unpooling operation~\cite{wu2024point} based on the voxel clustering induced in the symmetric downsampling stage.
The restored features $\tilde{\mathbf{X'}}_{g}^{\,l}$ are then fused with the aligned textual features $\mathbf{X'}_{t}^{L-l-1}$:
\begin{equation}
\label{eq:fusion}
\begin{aligned}
     \tilde{\mathbf{X}}_g^{l+1} &= \text{AttnBlock}(\text{SerialAttn}(\tilde{\mathbf{X'}}_{g}^{\,l}), \mathbf{X'}_{t}^{L-l-1}),
\end{aligned}
\end{equation}
where $\mathbf{X'}_{t}^{L-l-1}$ and $\tilde{\mathbf{X'}}_{g}^{\,l}$ share the aligned semantic level. 
The initial input $\tilde{\mathbf{X}}_{g}^{1}$ is set to the primitive point feature $\mathbf{X}_{g}^{L}$ from the final downsampling stage.

\subsubsection{Decoder}
\label{sec:appendix-decoder}

After upsampling, we obtain final primitive-line features $\tilde{\mathbf{X}}_g^{L} \in \mathbb{R}^{\sum_{i=1}^{|E_g|} n_g^{\,i} \times D}$, which are passed to the decoder for the final panoptic spotting prediction.

Specifically, the decoder first adopts group-wise pooling strategy~\cite{wei2026point} $\mathbf{F}_g = \text{Pool}(\tilde{\mathbf{X}}_g^{L})$ to aggregate line features within each primitive into primitive-level representations $\mathbf{F}_g \in \mathbb{R}^{N_g \times D}$ where $N_g=|E_g|$.

Then the decoder adopts layer enhancement operation~\cite{liu2024sympoint} through aggregating the primitives features within the same layer by max pooling, average pooling and attention pooling as the layer-wise context, which is added back get the enhanced primitive feature $\tilde{\mathbf{F}}_g = \mathbf{F}_g + \text{Pool}_{layer}(\mathbf{F}_g)$.

Next, we adopt a OneFormer3D~\cite{kolodiazhnyi2024oneformer3d} based head to obtain the final prediction. It composed $U$ layers and first inits the learnable queries $\mathbf{Q}^0 \in \mathbb{R}^{O \times D}$ by query selection where $O$ is the number of symbol queries. 

During the $u$-th layer within the decoder, the queries $\mathbf{Q}^u$ are refined through self-attention and cross-attention with primitive features $\tilde{\mathbf{F}}_g$ serves as key and value.
Final semantic and instance predictions is obtained via:
\begin{equation}
\label{eq:query-result}
\begin{aligned}
    \mathbf{I}^U &= f_{cls}(\mathbf{Q}^{U}), \\
\mathbf{M}^U &= f_{inst}(\mathbf{Q}^{U}) (\tilde{\mathbf{F}}_g)^\top,
\end{aligned}
\end{equation}
where $f_{cls}$ and $f_{inst}$ are \text{MLP} based heads producing the semantic output $\mathbf{I} =\mathbf{I}^U \in \mathbb{R}^{O \times C}$ and instance mask $\mathbf{M} = \mathbf{M}^U \in \mathbb{R}^{O\times N_g}$. $C$ is the number of categories. 
The decoder also applies post-processing strategy Branch Fusion Refinement~\cite{wei2026point} to resolve inconsistencies.

Finally, to align with the task formalization, we integrate $\mathbf{I}^U$ and $\mathbf{M}^U$ to produce the final panoptic spotting result for primitives:
\begin{equation}
\label{eq:final-result}
\begin{aligned}
   \mathbf{Y}, \mathbf{Z} = \Phi(\mathbf{I}^U, \mathbf{M}^U),
\end{aligned}
\end{equation}

\subsection{Loss function}
\label{sec:loss}

We apply the classification loss $\mathcal{L}_{cls}$, cross-entropy loss $\mathcal{L}_{sem}$, binary cross-entropy loss $\mathcal{L}_{bce}$, dice loss $\mathcal{L}_{dice}$ and to jointly optimize the task branches.
$\mathcal{L}_{cls}$ and $\mathcal{L}_{sem}$ supervise semantic predictions:
\begin{equation}
\label{eq:losses-semantic}
\begin{aligned}
\mathcal{L}_{cls} = \mathrm{CrossEntropy}(\mathbf{I}, \mathbf{I}^{gt}), 
\mathcal{L}_{sem} = \mathrm{CrossEntropy}(\mathbf{Y}, \mathbf{Y}^{gt}).
\end{aligned}
\end{equation}
$\mathbf{I},\mathbf{I}^{gt}\in\mathbb{R}^{O\times C}$ are the predicted and ground-truth classes of $O$ symbols over $C$ categories. $\mathbf{Y},\mathbf{Y}^{gt}\in\mathbb{R}^{N_g\times C}$ are the semantic predictions and labels over $N_g$ primitives. $\mathcal{L}_{bce}$ and $\mathcal{L}_{dice}$ supervise instance masks predictions:
\begin{equation}
\label{eq:losses-mask}
\begin{aligned}
 \mathcal{L}_{bce} = \mathrm{BinaryCrossEntropy}(\mathbf{M}, \mathbf{M}^{gt}), 
\mathcal{L}{dice} = 1 - \frac{2\langle \mathbf{M}, \mathbf{M}^{gt} \rangle}{\|\mathbf{M}\|_{1} + \|\mathbf{M}^{gt}\|_{1}}.
\end{aligned}
\end{equation}
$\mathbf{M},\mathbf{M}^{gt}\in\mathbb{R}^{O\times N_g}$ are predicted and ground-truth instance mask of $O$ symbols over $N_g$ primitives. $\langle \cdot, \cdot \rangle$ denotes the element-wise product summed over all elements, and $\|\cdot\|_1$ denotes the sum of all elements.


Besides task-specific losses, we apply complexity loss $\mathcal{L}_c$~\cite{louizos2018learning} to regularize semantic filtering process and promote sparsity in textual feature selection:
\begin{equation}
\label{eq:l0-loss}
\begin{aligned}
    \mathcal{L}_c = \sum_{l=1}^{L}\sum_{j=1}^{N_t} \sigma\!\left(r_j^l - \tau \log (\zeta - \gamma) \right),
\end{aligned}
\end{equation}
where $r_j^l \in \mathbf{r}^l$ denotes the semantic relevance of the 
$j$-th annotation at $l$-th filtering layer. $\gamma$ and $\zeta$ are predefined stretch limits. $\tau$ is the temperature and $\sigma(\cdot)$ is the \text{sigmoid} activation.
This term estimates the expectation of active gates across filtering layers, imposing an adaptive sparsity constraint.
Minimizing $\mathcal{L}_c$ encourages the model to retain those most semantically relevant textual features, yielding a more discriminative semantic filtering mechanism.

In our experiments, we set the loss weights $\lambda_{cls}:\lambda_{bce}:\lambda_{dice}:\lambda_{cls}:\lambda_{c}=2:5:5:5:0.0001$. The task-specific loss weight is set empirically according to~\cite{wei2026point}. 
Since $\mathcal{L}_c$ regularizes semantic filtering rather than directly supervising prediction, we assign it a small weight to balance regularization and task optimization. We evaluate $\lambda_c\in\{1,0.1,0.01,0.001,0.0001\}$ and find that $0.0001$ achieves the best performance and stability.

\section{Datasets, metrics and baselines}
\label{sec:dataset and metric}
Here we introduce additional details on datasets and evaluation metrics.

\subsection{Dataset} 
\label{sec:datasets} 
This section outlines the data preprocessing steps for two datasets.

\textbf{FloorPlanCAD-V2.}
We utilize the large-scale FloorPlanCAD-V2\cite{fan2021floorplancad} which is designed for panoptic symbol spotting.
Compared with its earlier version, this release offers improved scale and textual semantic diversity

FloorPlanCAD-V2 contains 35 line-level annotated categories, distinguishing between 30 countable “thing” classes (e.g. doors, windows) and 5 uncountable “stuff” classes (walls, curtain wall, parking spot, row chairs and railing).
We follow the official script\footnote{\url{https: //github.com/VITA-Group/CADTransformer/}} to assign semantic labels and instance indices to primitives, while additionally extracting relevant textual annotation information in paralle.
We split the dataset into train, validation and test sets with a approximate $\{6:3:1\}$ proportion, yielding $\{9533:4597:1533\}$.

\textbf{CubiCasa5K.}
To further evaluate the robustness of our method, we use CubiCasa5K which is a real-world CAD dataset for floorplan image analysis.

Since CubiCasa5K was not originally designed for this task, we refer to the official script\footnote{\url{https://github.com/CubiCasa/CubiCasa5k}} and the icon class mapping in the benchmark, obtaining 10 countable ``thing'' classes (window, door, closet, electrical appliance, toilet, sink, sauna bench, fireplace, bathtub and chimney) and 2 uncountable ``stuff'' classes (wall, railing), which serve as \textit{semantic labels} for primitives.
For \textit{instance indices}, all primitives belonging to the same symbol are assigned the same instance index for thing categories. For stuff categories, primitives are assigned only semantic labels and are not distinguished by instance indices.

We split CubiCasa5K into training, validation, and test sets with $\{4200:400:400\}$ samples, following the benchmark~\cite{kalervo2019cubicasa5k}.

\subsection{Evaluation metrics} 
\label{sec:metrics}
Following~\cite{fan2021floorplancad}, we use the comprehensive measurement \textbf{PQ} (Panoptic Quality), which is specifically designed for panoptic symbol spotting task, to jointly evaluate instance recognition and semantic segmentation:
\begin{equation}
\begin{aligned}
PQ = & \frac{|TP|}{|TP| + \frac{1}{2}|FP| + \frac{1}{2}|FN|} \times \frac{\sum_{(s_{pred}, s_{gt}) \in TP} \text{IoU}(s_{pred}, s_{gt})}{|TP|} \\
  = & \frac{\sum_{(s_{pred}, s_{gt}) \in TP} \text{IoU}(s_{pred}, s_{gt})}{|TP| + \frac{1}{2}|FP| + \frac{1}{2}|FN|},
\end{aligned}
\end{equation}
where a prediction symbol $s_{pred}$ (a set of primitives) is considered a match with ground truth $s_{gt}$ if they share the same predicted category and $\text{IoU}\allowbreak (s_{pred}, s_{gt})>0.5$. The intersection over union (IoU) score are computed as follows:
\begin{equation}
\begin{aligned}
\text{IoU}(s_{pred}, s_{gt}) = \frac{\sum_{e_g^i \in s_{pred} \cap s_{gt}} \log(1 + L(e_g^i))}{\sum_{e_g^j \in s_{pred} \cup s_{gt}} \log(1 + L(e_g^j))},
\end{aligned}
\end{equation}
where $e_g^i$ and $e_g^j$ denote graphical primitives and $L(\cdot)$ calculates the arc length.

We additionally report \textbf{PQ-Thing} and \textbf{PQ-Stuff} to measure the performance on countable thing and uncountable stuff categories, respectively. Semantic symbol spotting is evaluated using \textbf{F1} and length-weighted F1 score \textbf{wF1}, where longer primitives receive larger weights.

\subsection{Baselines} 
\label{sec:baselines}
We compare our model against ten baselines spanning multiple paradigms:
\begin{itemize}[leftmargin=*,itemsep=2pt, parsep=0pt]
\item \textbf{PanCADNet}~\cite{fan2021floorplancad} uses a CNN backbone for raster feature extraction, GCN for semantic detection, and Faster R-CNN for instance recognition.

\item \textbf{CADTransformer}~\cite{fan2022cadtransformer} represents primitives from rasterizd feature maps extracted by a CNN backbone and employs a modified Vision Transformer~\cite{dosovitskiy2020image} with neighborhood-attention mechinism to refine the features.
\item \textbf{GAT-CADNet}~\cite{zheng2022gat} constructs graph with primitives as nodes and employs GAT for feature propagation. It also establishes edge features to capture spatial relationships, enabling more effective message passing.

\item \textbf{SymPoint}~\cite{liu2024symbol} models graphical primitives as point cloud structure to enhance the feature
extraction. It uses Point Transformer to update features and employs Mask2Former decoder to obtain final predictions.
\item \textbf{SymPointV2}~\cite{liu2024sympoint} improves SymPoint by encoding layer assignments into primitive features and proposes a position-guide training method to accelerates the convergence of the model.
\item \textbf{DPSS}~\cite{luo2026archcad} follows the point cloud structure but incorporates raster features from images to enhance primitive features, improving the performance and the robustness of the model.

\item \textbf{VecFormer}~\cite{wei2026point} proposes line-based representation of graphical primitives, better preserving their original geometric structure. It also utilizes branch fusion refinement strategy to improve prediction reliability.

\item \textbf{PFL-Net}~\cite{wang2025panoramic} simply employs a pre-trained language model to embed text annotations, failing to capturing the semantics implicit in the complex syntactic structures of CAD annotations.
\item \textbf{TNet}~\cite{liu2026text} treats text annotations as a distinct primitive type and embeds them via feature maps, failing to exploit semantic value. 
\item \textbf{TriNet}~\cite{xing2026multimodal} relies on hand-crafted fusion of different modalities, where texts are merely assigned to graphical primitives by nearest-coordinate matching, without reasonable encoding and cross-modal alignment.
\end{itemize}

\section{Parameter sensitivity analysis}
\label{sec:param-analysis}
We conducted experiments to report three key hyperparameters.


\subsection{Embedding dimension}
TACE embeds text annotations into $\mathbf{X}_t^0 \in \mathbb{R}^{{N_t}\times{D}}$, where dimension $D$ controls the capacity to capture textual semantic. We evaluate different dimensions in~\tabref{tab:dim}. The best performance is achieved at $D=32$: smaller dimension limits the representation capacity, leading to insufficient semantic encoding, while a larger dimension introduces redundant information.

\begin{table}[htbp]
\centering
\caption{Embedding Dimension Selection.}
\label{tab:dim}
\renewcommand{\arraystretch}{0.9}
\resizebox{0.7\columnwidth}{!}{
\begin{tabular}{@{}l cccccc@{}}
\toprule
& Dim
& PQ
& PQ-Thing
& PQ-Stuff
& F1
& wF1 \\
\midrule
\multirow{3}{*}{\textbf{TACE}}
& 64 & 91.32 & 91.39 & 91.27 & 91.1 & 91.4 \\
& \textbf{32} & \textbf{91.69} & \textbf{91.84} & \textbf{91.54} & \textbf{91.7} & \textbf{91.6} \\
& 16 & 91.19 & 90.83 & 91.51 & 90.7 & 91.2 \\
\bottomrule
\end{tabular}
}
\end{table}

\subsection{Primitives downsampling layer} 
During Semantic Hierarchy Alignment, graphical primitive-lines are downsampled into higher-level representations.
We evaluate the number of downsampling layers of $\{7, 5, 3\}$ with grid size rate of $[1,2,2,1,2,1,2]$, $[1,2,2,2,2]$ and $[1,4,4]$ in~\tabref{tab:downsample-layer}.
Fewer layers provide insufficient hierarchical abstraction, while excessive downsampling layers may cause redundant transformations and feature smoothing. Five layers perform best.

\begin{table}[htbp]
\centering
\small
\caption{Downsampling Layer Selection.}
\label{tab:downsample-layer}
\renewcommand{\arraystretch}{0.9}
\resizebox{0.7\columnwidth}{!}{
\begin{tabular}{@{}ccccccc@{}}
\toprule
 & Layer & PQ & PQ-Thing & PQ-Stuff & F1 & wF1 \\
\midrule
\multirow{3}{*}{\textbf{Down}} 
 & 7 & 91.38 & 91.41 & 91.35 & 90.7 & 91.4 \\
 & \textbf{5} & \textbf{91.69} & \textbf{91.84} & \textbf{91.54} & \textbf{91.7} & \textbf{91.6} \\
 & 3 & 91.15 & 90.78 & 91.52 & 90.9 & 91.5 \\
\bottomrule
\end{tabular}
}
\end{table}

\subsection{Semantic filtering layer} 
During Multi-Level Semantic Filtering, textual features are progressively filtered to align with primitive features. The number of semantic filtering layers $L$ controls the hierarchical adaptivity of semantic alignment. 
Based on the parameter experiments of primitive downsampling layers, we evaluate different values of $L$ under five primitive downsampling layers. As shown in ~\tabref{tab:msf-layer}, $L=5$ achieves the best performance, indicating that retaining five semantic filtering layers best exploit adaptive semantic alignment.

\begin{table}[htbp]
\centering
\small
\caption{Semantic Filtering Layer Selection.}
\label{tab:msf-layer}
\renewcommand{\arraystretch}{0.9}
\resizebox{0.7\columnwidth}{!}{
\begin{tabular}{@{}ccccccc@{}}
\toprule
 & Layer & PQ & PQ-Thing & PQ-Stuff & F1 & wF1 \\
\midrule
\multirow{4}{*}{\textbf{MSF}}
& 0 & 91.50 & 90.83 & 92.08 & 90.9 & 91.4 \\
& 1 & 91.77 & 91.81 & 91.73 & 91.2 & 91.7 \\
& 3 & 91.97 & 92.15 & 91.80 & 91.8 & 91.5 \\
& \textbf{5} & \textbf{92.67} & \textbf{93.05} & \textbf{92.32} & \textbf{93.4} & \textbf{91.9} \\
\bottomrule
\end{tabular}
}
\end{table}


\begin{figure*}[htbp]
    \centering
    \begin{subfigure}[b]{0.31\textwidth}
        \centering
        \includegraphics[width=\linewidth]{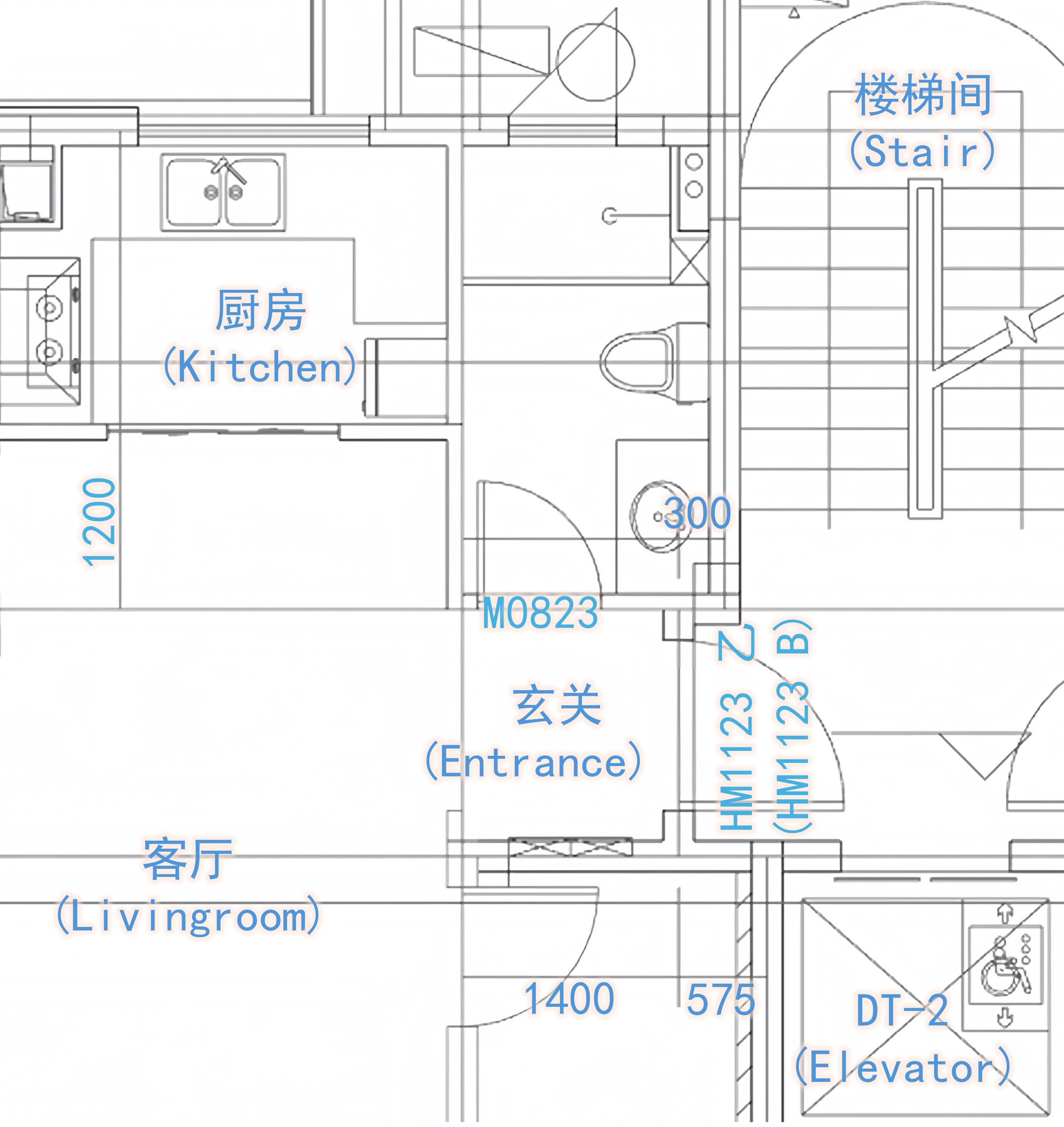}
        \label{sub-fig:layer1}
    \end{subfigure}
    \hspace{2pt}\vrule width 0.2pt\hspace{2pt}
    \begin{subfigure}[b]{0.31\textwidth}
        \centering
        \includegraphics[width=\linewidth]{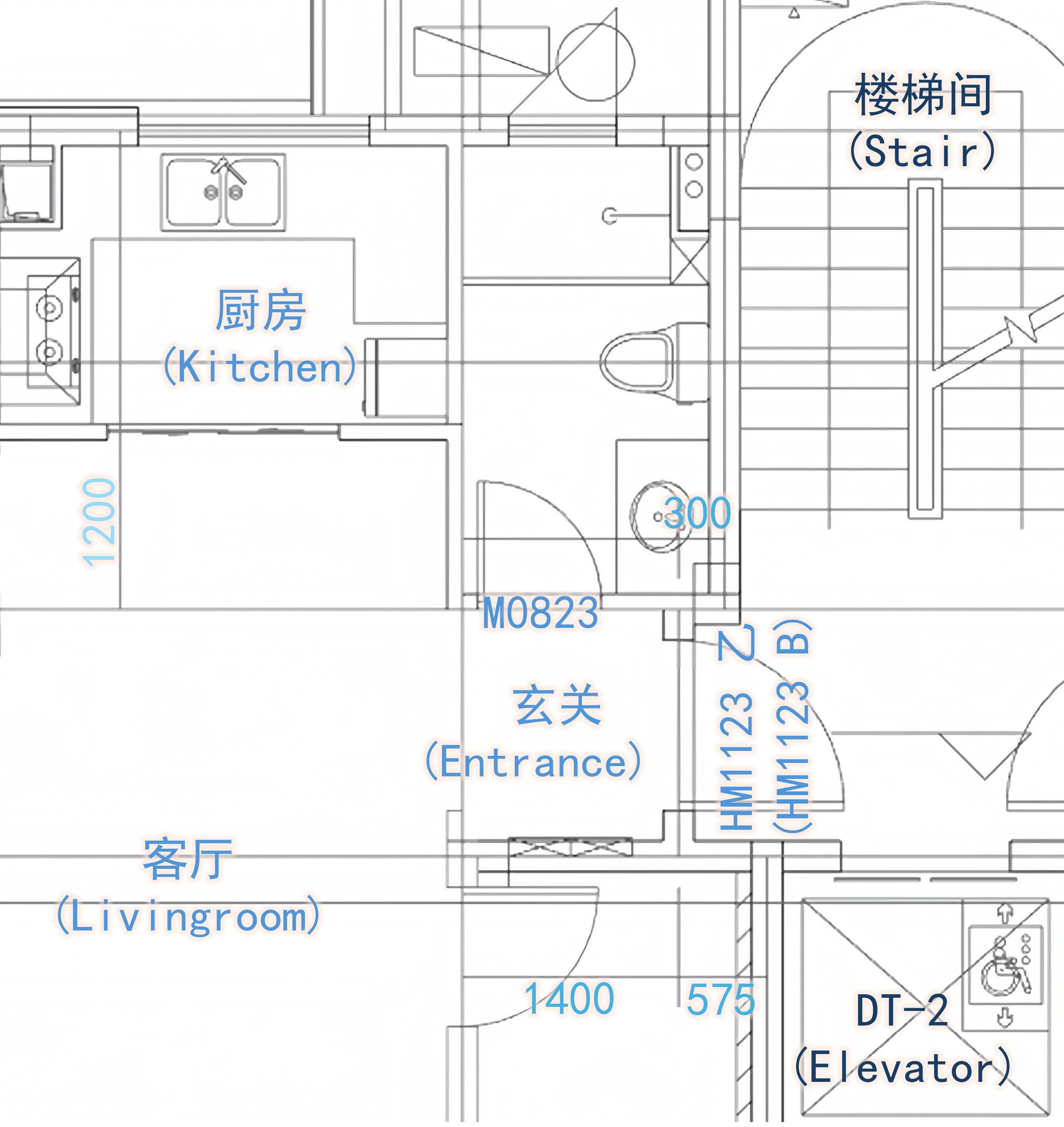}
        \label{sub-fig:layer2}
    \end{subfigure}
    \hspace{2pt}\vrule width 0.2pt\hspace{2pt}
    \begin{subfigure}[b]{0.31\textwidth}
        \centering
        \includegraphics[width=\linewidth]{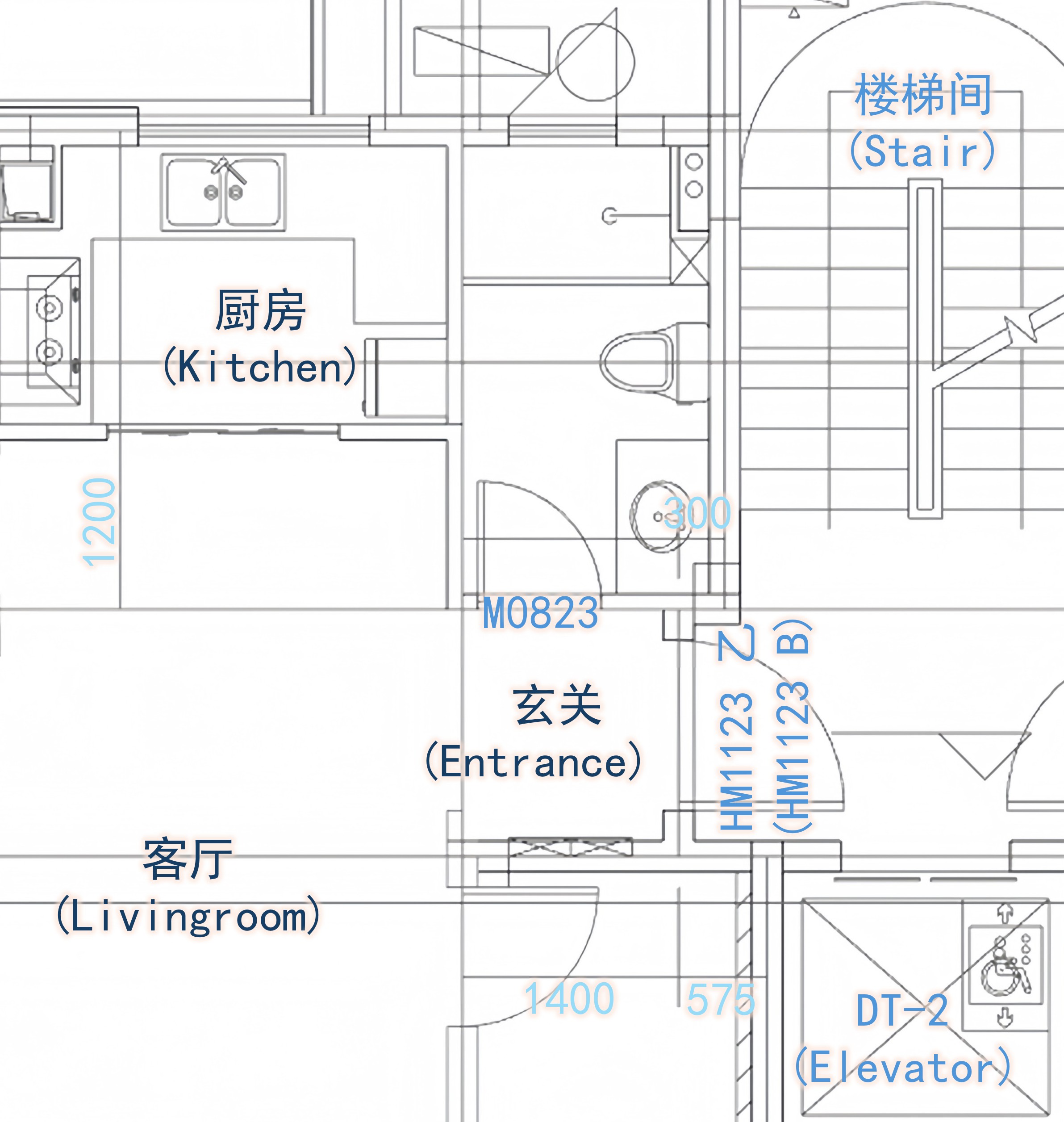}
        \label{sub-fig:layer3}
    \end{subfigure}

    \caption{Case study of MSF.}
    \label{fig:msf-case-study}
\end{figure*}

\section{Additional case study}
\label{sec:msf-case-study}
We further illustrate the effectiveness of our proposed MSF module \allowbreak through an interesting case study.

~\figref{fig:msf-case-study} illustrates the semantic filtering results of the third, forth and fifth layers within MSF, shown from left to right (in our configuration, the frist layer performs no actual filtering but only projection to align dimensions, while the second layer mainly serves as a transitional stage where the semantic hierarchy has not yet been clearly differentiated). 
We visualize the retention degree of each textual annotation at layer $l$ according to the approximated binary gate $\mathbf{G}^l$: darker colors indicate higher retention gate (closer to 1), and lighter colors indicate lower retention gate (closer to 0).

As shown in ~\figref{fig:msf-case-study}, filtering results are largely consistent with the semantic hierarchy. Primitive-level annotations such as `\texttt{300}' exhibit gradually weakened semantics as filtering deepens; semantics of instance-level annotations such as `\texttt{DT-2}' remain highly preserved in intermediate layers; region-level annotations such as `\texttt{Entrance}' tend to be increasingly retained in higher layers. 
In parallel, downsampling naturally produces higher-level primitive feature abstractions, which is a standard property, resulting in aligned semantic across modalities.
Note that the retention gate of instance or region-level is generally higher than primitive-level, probably because higher-level annotations convey more informative and semantically richer cues.

\section{Additional qualitative results}
\label{sec:qualitative-result}
\begin{figure}[htbp]
    \centering
    \begin{subfigure}[b]{0.185\linewidth}
        \centering
        \includegraphics[width=\linewidth]{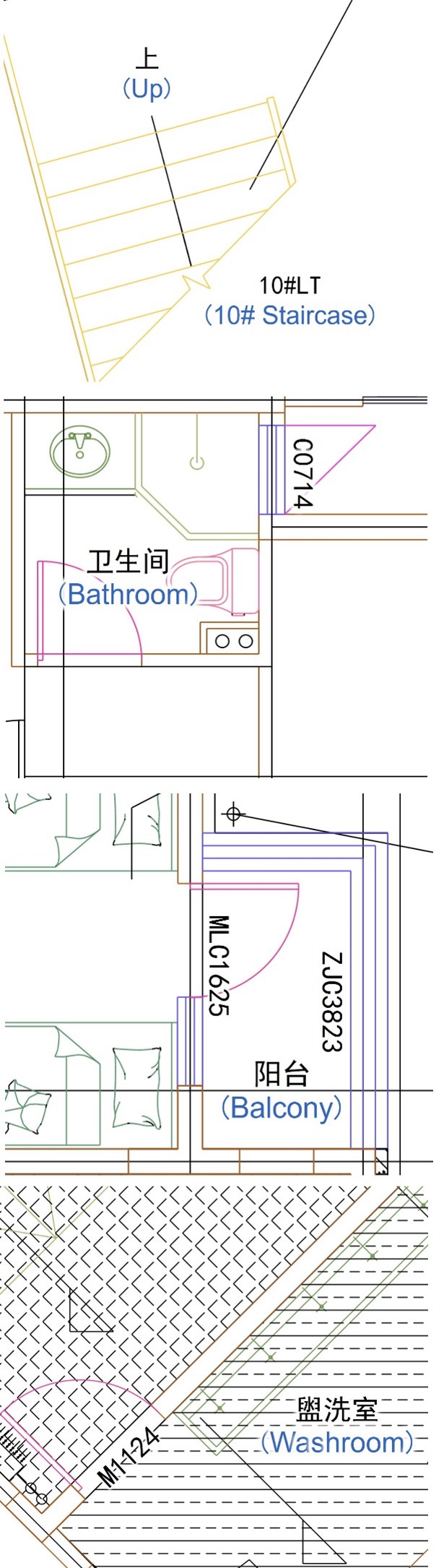}
        \caption{GT}
        \label{sub-fig:GT-app}
    \end{subfigure}%
    \hfill
    \vrule width 0.1pt
    \hfill
    \begin{subfigure}[b]{0.185\linewidth}
        \centering
        \includegraphics[width=\linewidth]{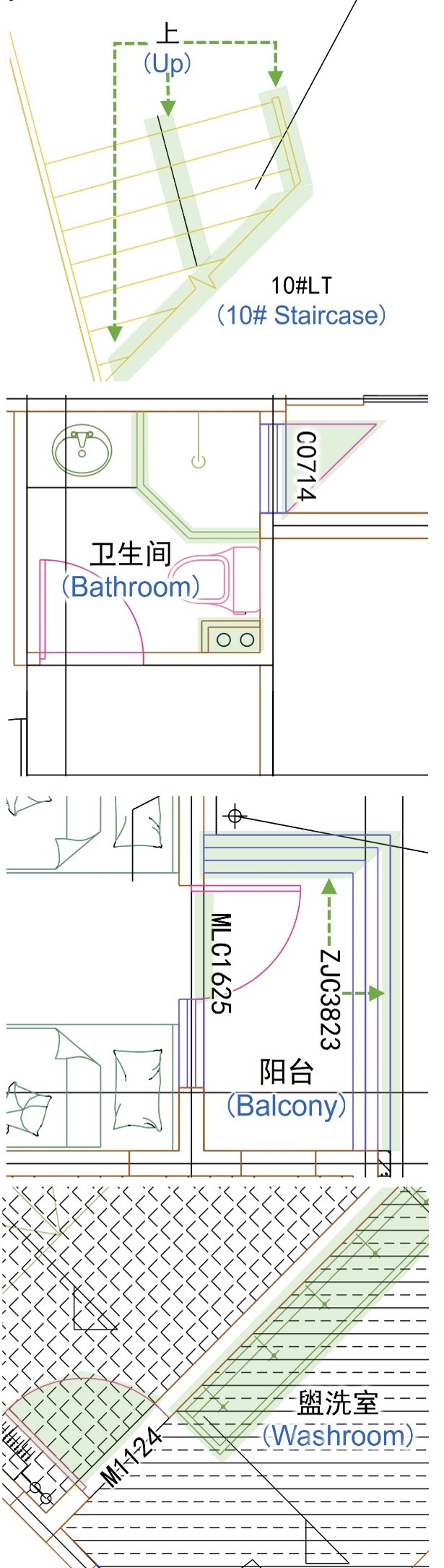}
        \caption{TextCAD}
        \label{sub-fig:TextCAD-app}
    \end{subfigure}%
    \hfill
    \vrule width 0.1pt
    \hfill
    \begin{subfigure}[b]{0.185\linewidth}
        \centering
        \includegraphics[width=\linewidth]{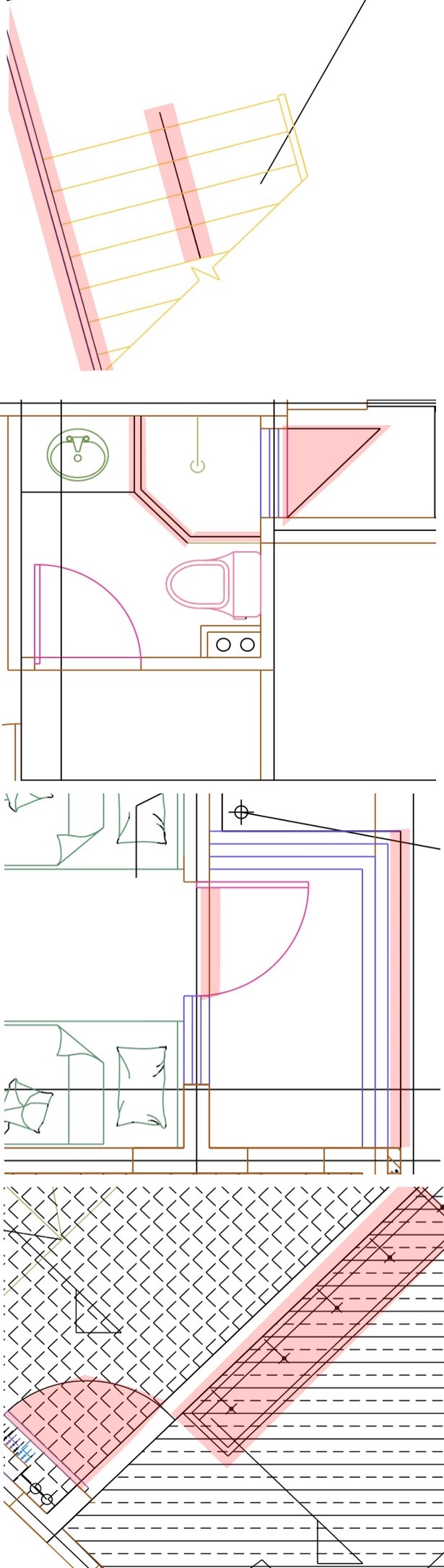}
        \caption{CADTrans.}
        \label{sub-fig:CADTransformer-app}
    \end{subfigure}%
    \hfill
    \vrule width 0.1pt
    \hfill
    \begin{subfigure}[b]{0.185\linewidth}
        \centering
        \includegraphics[width=\linewidth]{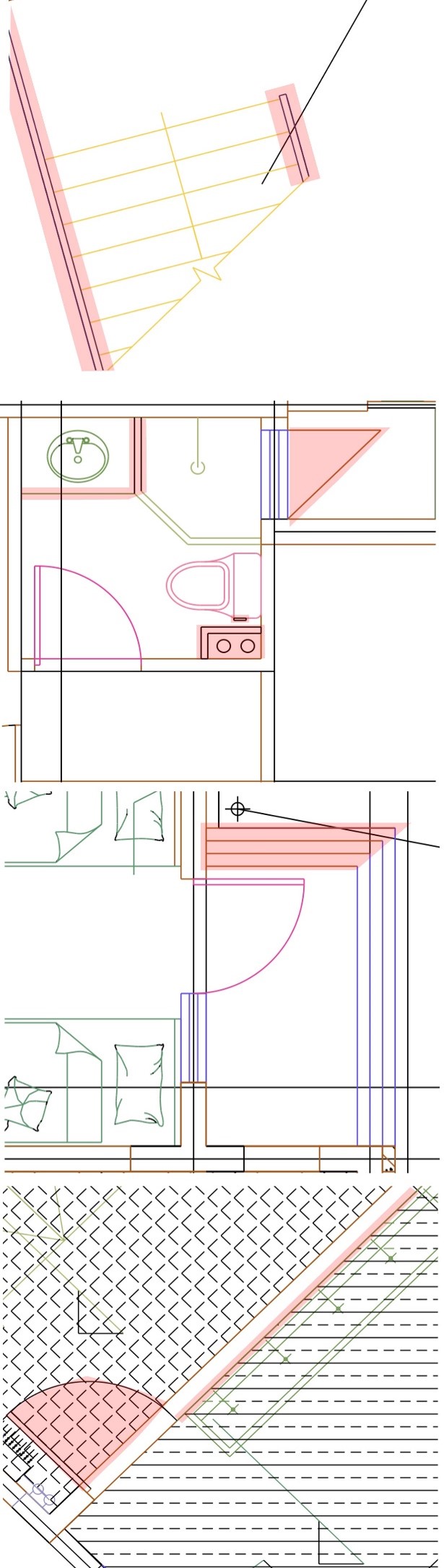}
        \caption{SymPoint}
        \label{sub-fig:SymPoint-app}
    \end{subfigure}%
    \hfill
    \vrule width 0.1pt
    \hfill
    \begin{subfigure}[b]{0.185\linewidth}
        \centering
        \includegraphics[width=\linewidth]{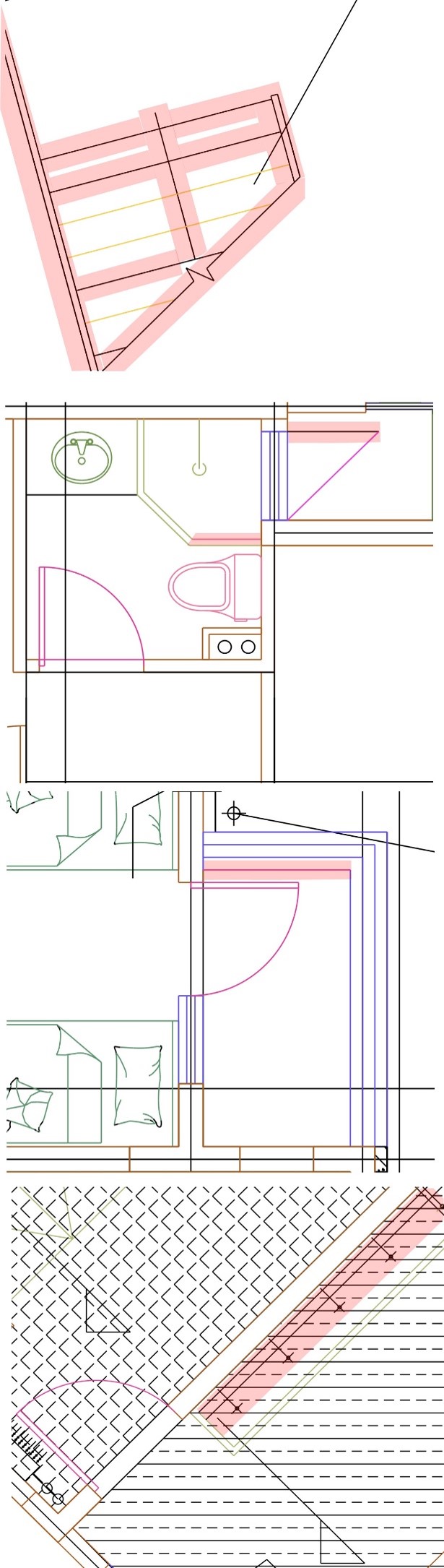}
        \caption{VecFormer}
        \label{sub-fig:VecFormer-app}
    \end{subfigure}

    \caption{Additional qualitative comparison among different methods.}
    \label{fig:additional-qualitative-comparison}
\end{figure}

Additional qualitative results are shown in~\figref{fig:additional-qualitative-comparison}.
TextCAD achieves more accurate predictions in challenging regions across diverse scenarios by leveraging underlying semantic cues from relevant textual annotations.


\clearpage
\bibliographystyle{elsarticle-num}
\bibliography{reference}

\end{document}